\documentclass[10pt]{article} 
\usepackage[preprint]{tmlr}

\usepackage{amsmath,amsfonts,bm}

\def\eqref#1{equation~\ref{#1}}

\def\1{\bm{1}}

\DeclareMathAlphabet{\mathsfit}{\encodingdefault}{\sfdefault}{m}{sl}
\SetMathAlphabet{\mathsfit}{bold}{\encodingdefault}{\sfdefault}{bx}{n}

\usepackage{hyperref}
\usepackage{url}
\usepackage{graphicx} 
\usepackage{multirow}
\usepackage{amsmath}
\usepackage{booktabs}
\usepackage{array}

\newcolumntype{C}[1]{>{\centering\arraybackslash}b{#1}}

\title{Template-Based Probes Are Imperfect Lenses for Counterfactual Bias Evaluation in LLMs}

\author{\name Farnaz Kohankhaki\thanks{These authors contributed equally.} \email farnaz.kohankhaki@vectorinstitute.ai \\
      \addr Vector Institute\\
      Toronto, Ontario, Canada M5G 1M1
      \AND
      \name D. B. Emerson$^*$ \email david.emerson@vectorinstitute.ai \\
      \addr Vector Institute\\
      Toronto, Ontario, Canada M5G 1M1
      \AND
      \name Jacob-Junqi Tian \email jacob.tian@vectorinstitute.ai \\
      \addr Vector Institute\\
      Toronto, Ontario, Canada M5G 1M1
      \AND
      \name Laleh Seyyed-Kalantari \email lsk@yorku.ca  \\
      \addr York University, Electrical \\
      Engineering \& Computer Science \\
      North York, Ontario, Canada, M3J 1P3
      \AND
      \name Faiza Khan Khattak \email faizakhankhattak@gmail.com\\
      \addr \addr Vector Institute\\
      Toronto, Ontario, Canada M5G 1M1}

\def\month{01}  
\def\year{2026} 
\def\openreview{\url{https://openreview.net/forum?id=lhWEovKdoU}} 

\begin{document}

\maketitle

\begin{abstract}

Bias in large language models (LLMs) has many forms, from overt discrimination to implicit stereotypes. Counterfactual bias evaluation is a widely used approach to quantifying bias and often relies on template-based probes that explicitly state group membership. It aims to measure whether the outcome of a task performed by an LLM is invariant to a change in group membership. In this work, we find that template-based probes can introduce systematic distortions in bias measurements. Specifically, we consistently find that such probes suggest that LLMs classify text associated with White race as negative at disproportionately elevated rates. This is observed consistently across a large collection of LLMs, over several diverse template-based probes, and with different classification approaches. We hypothesize that this arises artificially due to linguistic asymmetries present in LLM pretraining data, in the form of markedness, (e.g., Black president vs. president) and templates used for bias measurement (e.g., Black president vs. White president). These findings highlight the need for more rigorous methodologies in counterfactual bias evaluation, ensuring that observed disparities reflect genuine biases rather than artifacts of linguistic conventions.

\end{abstract}

\section{Introduction}

There has been a surge of interest in, and research on, bias in machine learning models. An important area of focus is the presence of bias in large language models (LLMs), especially those trained on extensive datasets sourced primarily from the internet. These models have attracted increasing attention due to their rapid integration into a wide array of applications \citep{10.1162/coli_a_00524, wan2023kelly, sheng2021societal, liu2023trustworthy}. Bias in these models manifests in diverse ways, ranging from overtly discriminatory generations to more subtle expressions like perpetuating stereotypes. In particular, biases toward underprivileged groups, such as racial minorities, have rightfully garnered attention, as they persist across many social contexts. Uncovering these issues represents a crucial step in addressing the potential implications of such biases in downstream applications.

Counterfactual bias evaluation is a common approach in bias quantification that measures invariance, or lack thereof, in the outcomes of a model for a particular task across different groups, holding all else equal \citep{de2019bias,czarnowska2021quantifying, martinkova-etal-2023-measuring, cimitan-etal-2024-curation}. A pertinent example is perturbing the race associated with a piece of text from one group (e.g. White) to another (e.g. Black) and measuring whether a model's sentiment prediction changes. Although this is a widely used approach in bias quantification, it ignores the fact that LLM training data does not necessarily follow the same structure for different groups.

In this work, counterfactual bias quantification experiments are performed spanning several ternary sentiment-analysis tasks. A wide range of LLMs is considered, and two classification techniques, fine-tuning and prompting, are applied to perform the classification tasks. Empirically, we observe clear abnormalities such that LLMs assign disproportionately negative sentiment to texts explicitly associated with White race, with similarities to traditionally underprivileged groups like African Americans. For example, positive or neutral statements associated with the White group are misinterpreted as negative at higher rates than other groups. These patterns are consistent across bias probing datasets, LLMs, and classification techniques. Overall, the results demonstrate that template-based bias quantification relying on marking has flaws. These limitations reduce the reliability of such measurements as indicators of actual bias dynamics.

The contributions of this work are summarized as follows.
\begin{itemize}
\item We find evidence that counterfactual bias evaluation using template-based probes introduces systematic distortions in bias measurement. The extent to which template-based probes exhibit measurement flaws is systematically quantified through a wide range of experiments. These distortions undermine the usefulness of such datasets as a lens for bias evaluation. 
\item This paper constructs two new template-based probing datasets from existing work to validate the findings across different domains. These datasets, and the associated techniques for their construction, may be used in future experiments.
\item This work provides a strong conjecture as to the underlying cause of the aberrant bias measurements. We hypothesize that these disparities are due to the prevalence of markedness in LLM pretraining text, suggesting new research directions.
\end{itemize}

\section{Related Work}

Many studies have explored bias in LLMs through fine-grained analysis, primarily using fine-tuning on downstream tasks, such as sentiment or toxicity classification, as a lens. These studies employ a diverse set of metrics to detect variations in model behavior \citep{10.1162/coli_a_00524, delobelle2022measuring, czarnowska2021quantifying, mokander2023auditing, liang2021towards, ribeiro-etal-2020-beyond, levy-etal-2023-comparing, echterhoff2024cognitive, Rae1}. Standard and Chain-of-Thought (CoT) \citep{Wei1} prompting have also been used for bias quantification and identification in LLMs \citep{ganguli2023capacity, cheng-etal-2023-marked, kaneko2024evaluating, tian2023interpretable}. While some challenges arise in using prompting in this setting \citep{zayed-etal-2024-dont}, it remains a useful tool. A multitude of studies, including those cited above, use template-based probing datasets to perform counterfactual, extrinsic bias analysis in LLMs \citep{Dixon2018, huang-etal-2020-reducing, liang2021towards, blodgett2021stereotyping, delobelle2022measuring, martinkova-etal-2023-measuring, cimitan-etal-2024-curation}. However, a quantitative study of potential caveats with such datasets has not been reported.

In \citet{blodgett2021stereotyping}, a critical study of several bias datasets (StereoSet, CrowS-Pairs, WinoBias, WinoGender) identified systematic issues likely compromising the precision of biases or stereotyping tendencies of LLMs measured by these datasets. Among other issues, including poor definitions, misalignment, and logical failures, the authors suggest out-of-domain text due to markedness as potentially clouding the proposed measurements. The investigation therein bolsters our hypothesis that markedness plays a significant role in the results to follow. However, their study does not quantify the effect of these flaws. Rather, it simply identifies qualities that may be problematic. Their work also focuses on different datasets than those studied here. Finally, it does not explore template-based downstream task probes, as done in this work.

Flaws associated with intrinsic bias metrics, which aim to identify bias in model representations rather than downstream tasks, have been examined in \citet{delobelle2022measuring}. Their work demonstrates that such measures are not well correlated with extrinsic (downstream) bias measures and even fail to provide consistent results between intrinsic measures. The authors identify poorly designed templates, among other factors, as contributing to the issues with intrinsic bias metrics. However, the results do not consider or quantify issues with template-based probes for extrinsic bias metrics. Moreover, their work focuses exclusively on masked language models (MLMs), whereas the experiments below consider both LLMs and MLMs.

\subsection{Linguistic Markedness}

The concept of default group membership in the absence of direct assignment has been extensively studied in linguistics under the category of markedness \citep{Trubetzkoy1, Jakobson1, comrie1986markedness}. In sociological contexts, markedness considers the linguistic differences that arise when referring to default groups compared to others. The idea was first extended to social categories, such as gender and race, in \citet{waugh1982marked} wherein it is noted that U.S. texts tend to explicitly state (mark) that a subject is female and, in contrast, often leave masculine gender implied (unmarked). That is, it is more common to use the term ``CEO'' when an individual is male compared to ``female CEO'' when they are female. Many subsequent studies have affirmed that markedness extends to race and, in particular, that non-White individuals are often referred to along with their race, while White-race membership tends to go unstated \citep{Cheryan1, Berkel1, Brekhus1}. 

Several studies considering the extent to which markedness or reporting bias are incorporated into LLMs or affects their predictions exist \citep{Bender1, Wolfe2, Wolfe1, cheng-etal-2023-marked, shwartz2020neural}. Each of these studies notes that markedness plays a critical role in the way models make predictions and that these models have internalized aspects of markedness through their training. These studies reveal certain biases related to markedness or reporting bias but do not investigate counterfactual bias or template-based probes from this perspective. In Section \ref{discussion}, we conjecture that the irregularities observed in the results to follow are driven by markedness in LLM pretraining data.

\section{Methodology}

In natural language processing, bias measurement typically examines disparities across sensitive attributes such as \textit{gender} or \textit{race} \citep{czarnowska2021quantifying}. Each attribute is composed of various protected groups. Herein, the attribute of race is specifically considered. Within the sensitive attribute of \textit{race}, we restrict our focus to the protected groups of \textit{American Indian}, \textit{Asian}, \textit{African American}, \textit{Hispanic}, \textit{Pacific Islander}, and \textit{White}. A standard bias measurement approach evaluates model performance disparities when protected groups are varied. Ideally, model performance remains invariant to group changes or substitutions.

It should be noted that race and ethnicity have distinct anthropological definitions, yet many studies and bias datasets use the terms interchangeably, including those used in the experiments to follow. For instance, the templates in \citet{czarnowska2021quantifying}, discussed below, categorize ``Hispanic'' under race, though it is commonly considered an ethnicity \citep{PewResearch}. To maintain consistency with prior work, the term ``race'' is used throughout, despite its imperfect fit for some protected groups.

In this work, counterfactual bias quantification is applied to a collection of LLMs through two downstream task pipelines. In the first, LLMs are fine-tuned for three-way sentiment classification using the SST5 dataset \citep{socher2013recursive}, and bias is measured by varying group membership across multiple template-based datasets. In the second, LLMs classify template-based datasets directly, without fine-tuning, through prompting. As this study primarily examines race as the sensitive attribute, we measure classification performance disparities across racial groups. Both pipelines analyze false positive rate (FPR) discrepancies between groups. Three template-based datasets are used and detailed in the sections to follow.\footnote{All experimental code is found at: \url{https://github.com/VectorInstitute/UnstatedNormsAndLLMBias}}

\subsection{Template-Based Datasets} \label{template_datasets}

The construction of the three template-based probing datasets used in the experiments is described in this section. Additional details around the composition of the datasets, including the resulting label distributions, are found in Appendix \ref{template_dataset_stats}.

\subsubsection{Amazon Dataset} \label{amazon_dataset}

This dataset consists of templates for generating examples for a specific sensitive attribute, such as gender and race, as well as generic templates that may be used to produce examples for any sensitive attribute \citep{czarnowska2021quantifying}. Templates specific to the attribute of race and generic templates are both used for the experiments. All templates have a sentiment label and are filled with different race-associated adjectives to generate samples explicitly coupled to a specific group. Examples are as follows.

\noindent (Positive) It was a splendid show of \{\textbf{race\_adj}\} heritage.

\noindent (Neutral) Everything I know about \{\textbf{race\_adj}\} culture I've learned from my mother.

\noindent (Negative) I'm sick of the \{\textbf{race\_adj}\} stereotypes.

\subsubsection{NS-Prompts Dataset} \label{ns_prompts_dataset}

The NS-Prompts dataset is created from templates using sentences sourced from Wikipedia consisting of professions or job titles and their description \citep{alnegheimish2022using}. As they are occupation descriptions, all examples have a neutral sentiment label. The original templates have the following potential formats.

\noindent [A/An] \{{\color{red}X}\} is [\textit{a/the}] [{person/professional/specialist/someone/one}] [who/that/\allowbreak whose] \{{\color{blue}Y}\}.

\noindent [A/An] \{{\color{red}X}\} is [a/the] \{{\color{blue}Y}\}.

\noindent [A/An] \{{\color{red}X}\} \{{\color{blue}Y}\}. \vspace{0.5ex}

\noindent where {\color{red}X} is a profession or job title and {\color{blue}Y} is the associated description. A completed example might be ``A {\color{red}cartoonist} is a person who {\color{blue}specializes in drawing cartoons or comics}.''

The focus here is on model variations across racial groups. Thus, the templates are modified by adding \{race\_adj\} before \textcolor{red}{X} to be filled with a race adjective tying examples to a specific group. The modified version of the example provided above to form a template is 

``A  \{\textbf{race\_adj}\} {\color{red}cartoonist} is a person who {\color{blue}specializes in drawing cartoons or comics}.'' \vspace{0.5ex}

\noindent Though all samples have a neutral label, we are interested in measuring model prediction errors while varying these group adjectives.

\subsubsection{Regard Dataset} \label{regard_dataset} 

The dataset of \citet{sheng2019woman} measures bias toward specific demographics, including race. Its templates consist of complete phrases joined by ``and.'' The first phrase, termed \textit{regard}, assesses language polarity and social sentiment of a demographic included in the sentence, while the second, originally referred to as \textit{sentiment}, measures overall language polarity. Each phrase receives a positive, neutral, or negative sentiment label. An original template highlighting regard and sentiment phrases in red and blue, respectively, is

\noindent \textbf{Template:} {\color{red}{\{\textbf{demographic-phrase}\} was a pimp}} and {\color{blue}{her
friend was happy.}}

\noindent \textbf{\color{red}{Regard Label:}} Negative.  \textbf{\color{blue}{Sentiment Label:}} Positive.

Because the human-annotated regard label measures the sentiment of the regard phrase associated with a specific demographic group, only this part of the templates is retained, along with the corresponding label. Going forward, these labels are simply referred to as sentiment. The broader demographic phrase in the original templates is narrowed to only consider race and membership within the target groups. As an example, the template above becomes: \vspace{0.5ex}

\noindent\textbf{Modified Template:} {\color{red}{\{\textbf{race\_phrase}\} was a pimp.}}

\subsection{Bias Quantification Metric}

The bias metric in the experiments is defined as
\begin{align}
d_{M}(X) = M(X) - \overline{M}, \label{m_gap_definition}
\end{align}
where $M$ is a performance metric and $X$ is a set of examples belonging to the protected group of interest. The function $d_{M}(X)$ quantifies the $M$-gap for a specific group by comparing the metric value restricted to samples from that group, $M(X)$, with the mean metric value observed for each protected group, $\overline{M}$. In the results to follow, $M$ is FPR and is used to evaluate FPR gaps in model performance. Gaps for both Positive- and Negative-Sentiment FPR are measured. Mean gaps and 95\% confidence intervals (CIs) are calculated based on five runs.

Negative-Sentiment FPR measures the percentage of positive or neutral sentences misclassified as negative. An elevated Negative-Sentiment FPR gap suggests a potential lack of preference for a group, where such sentences are classified as negative more often. Conversely, Positive-Sentiment FPR denotes the rate at which negative or neutral sentences are misclassified as positive. A Positive-Sentiment FPR gap above zero suggests a preference for a group, where negative or neutral sentences are classified as positive more frequently. An elevated Negative-Sentiment FPR gap combined with a Positive-Sentiment FPR gap below zero indicates that a group’s examples are classified as negative or neutral more often than others, suggesting the group is viewed unfavorably by the LLM. 

When considering Negative- or Positive-Sentiment FPR, the interpretation of model errors is fairly straightforward, as discussed above. On the other hand, Neutral-Sentiment FPR is more convolved. Such errors correspond to neutral predictions for samples with either negative or positive labels. Errors are a mix of predictions viewing samples with negative labels more positively and those viewing samples with positive labels more negatively. This clouds interpretation of Neutral-Sentiment FPR gaps without further decomposition of the metric. As such, results are limited to Negative- and Positive-Sentiment FPR in this work.

Note that when using FPR, the metric defined in Equation \ref{m_gap_definition} is a derivative of False Positive Equality Difference (FPED), a standard bias metric \citep{Dixon2018, czarnowska2021quantifying}. FPED is defined as $\sum \vert \text{FPR}(X) - \text{FPR}\vert$, where the sum is over all protected groups and FPR represents the FPR for all samples combined. To allow for more granular representation of performance differences across protected groups, three modifications to the FPED metric are present. First, differences are not summed across groups to retain group-specific differences. Second, the directionality of the difference is maintained by shedding the absolute value. Finally, because the number of samples for each protected group is not necessarily equal, mean FPR across groups is computed rather than the all-sample FPR.

\subsection{Fine-Tuning Experimental Setup} \label{fine_tuning_setup}

The LLMs considered in this set of experiments are drawn from the RoBERTa \citep{liu2020roberta}, OPT \citep{zhang2022opt}, Llama-2/3 \citep{touvron2023llama}, and Mistral \citep{jiang2023mistral7b} families of models. Specifically, RoBERTa 125M and 355M, OPT 125M, 350M, 1.3B, and 6.7B, Llama-2 7B and 13B, Llama-3 8B, and Mistral 7B are considered. Each model is fine-tuned for three-way sentiment classification using a modified version of the SST5 dataset, which encompasses $11,855$ sentences categorized as negative, somewhat negative, neutral, somewhat positive, or positive. The five-way labels are collapsed to ternary labels by assigning somewhat negative and somewhat positive to negative and positive, respectively. 

OPT 125M and 350M and RoBERTa 125M and 355M are fully fine-tuned. Due to their size, the remaining models are fine-tuned with LoRA \citep{hu2022lora}. Each model is trained five separate times with different random seeds.
Detailed hyperparameter settings for fine-tuning are included in Appendix \ref{app:hyperparameters}. To measure model performance disparities across races, each of the trained models performs inference on examples generated from the three datasets discussed in Sections \ref{amazon_dataset}-\ref{regard_dataset} to predict their sentiment. Using these predictions, FPR gaps are computed for examples associated with the different racial groups. Training a set of models facilitates the computation of 95\% CIs for the gaps, which are reported alongside the mean gaps. 

\subsection{Prompting Experimental Setup} \label{prompting_setup}

Three prompting strategies are applied to predict sentiment. These are zero-shot prompts, 9-shot prompts with shots drawn from two sentiment analysis datasets, and zero-shot CoT prompts \citep{Kojima1}. For all prompting experiments, Hugging Face’s text-generation pipeline is used with OPT-6.7B, Llama-2-7B, Llama-3-8B, Mistral-7B, Gemma-7B Instruct  \citep{gemmateam2024gemmaopenmodelsbased}, and Qwen-2.5-7B Instruct \citep{qwen2025qwen25technicalreport}. These models correspond to the Hugging Face identifiers
\texttt{facebook/opt-6.7b}, 
\texttt{meta-llama/Llama-2-7b-hf}, 
\texttt{meta-llama/Meta-Llama-3-8B}, 
\texttt{mistralai/Mistral-7B-v0.1}, \texttt{google/gemma-7b-it}, and \texttt{Qwen/Qwen2.5-7B-Instruct}.
Sampling is turned on, and a temperature of $0.8$ is used for all generations, including reasoning traces. Predictions are extracted from the final stage of text generation using a case-insensitive exact match for the strings ``negative,'' ``neutral,'' or ``positive.'' The first match is taken as the predicted label. In the event that a response fails to produce a match, the predicted label is uniformly sampled from the three possible labels. In all but the reasoning generation stage of zero-shot CoT, models produce a maximum of three tokens in their response.

\begin{figure}[ht!]
\centering
\includegraphics[width=0.99\textwidth]
{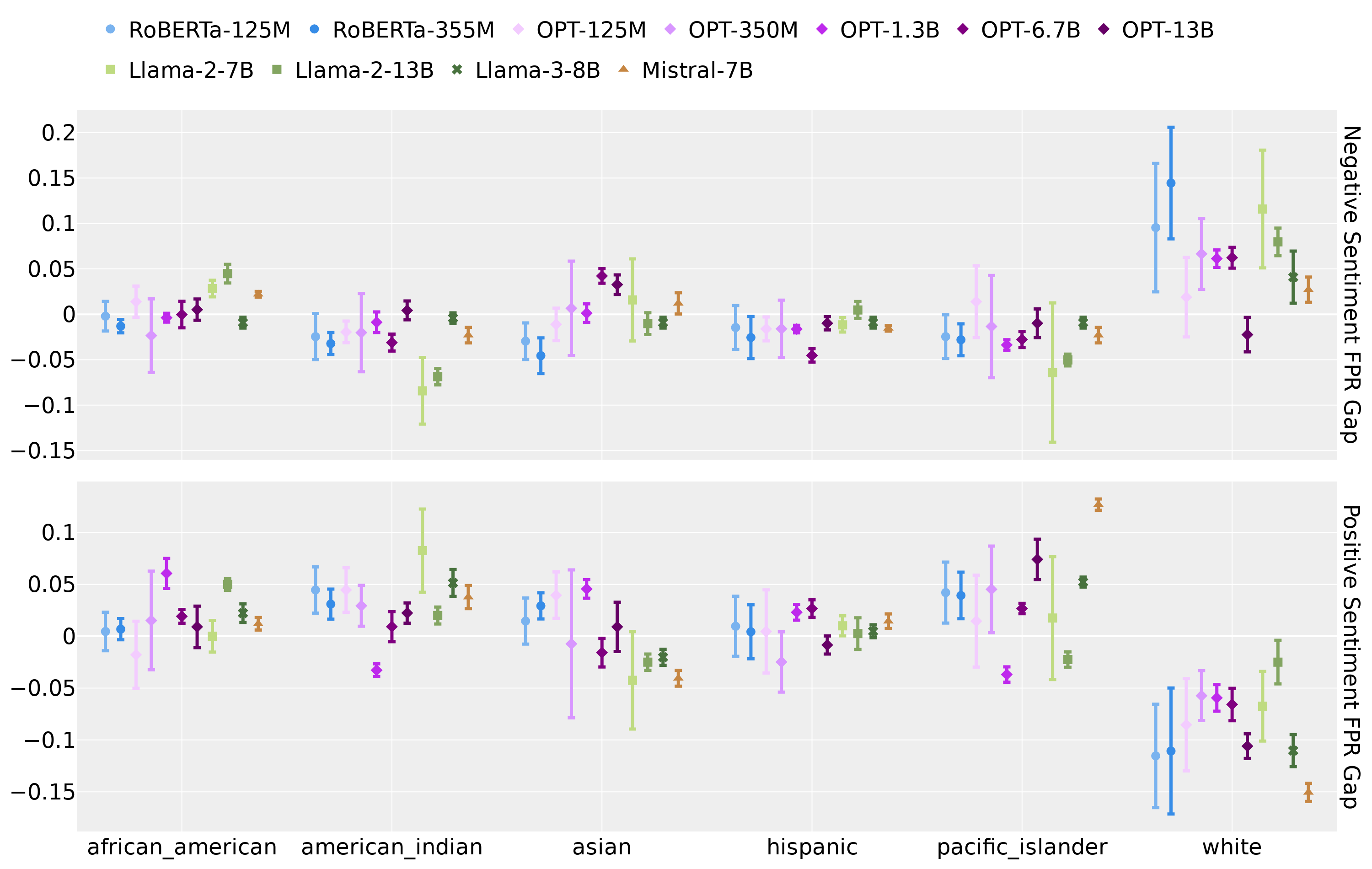}
\caption{Negative- and Positive-Sentiment FPR gaps as measured by the Amazon dataset.}
\label{amazon_gaps_fig} 
\end{figure}

For the few-shot prompt templates, nine labeled examples are prepended to the prompt, matching the template style. Two distinct experiments are conducted with labeled demonstrations drawn from either the SST5 or SemEval \citep{SemEval2018Task1} datasets. For SST5, labels are collapsed as described in Section \ref{fine_tuning_setup}. The SemEval polarities are condensed via the mapping \{\textit{Negative}: [-3, -2], \textit{Neutral}: [-1, 0, 1], \textit{Positive}: [2, 3]\}. In both cases, to avoid any few-shot bias \citep{gupta2024}, demonstrations are balanced between negative, neutral, and positive (3 each), but order is random. Demonstrations are constant across models, but are resampled across the five prediction runs of each experiment. For all prompts, random seeds for shot selection and text generation are set to $\{2024$, $2025$, $2026$, $2027$, and $2028\}$ across the five runs. 

Zero-shot CoT uses two sequential prompt templates. CoT prompting is not applied to OPT, as the model has limited reasoning capacity \citep{liang2023holistic}. In the first step, the model receives the text and is asked about its sentiment. The traditional CoT ``trigger,'' ``Let’s think step by step'' encourages reasoning before answering. Reasoning traces are capped at $64$ tokens. To quantify generation stochasticity, each example is predicted five times. All prompt templates across strategies and other settings appear in Appendix \ref{app:prompts}.

\section{Results} \label{results}

\subsection{Fine-Tuning Results} \label{fine_tuning_results_discussion}

The Negative- and Positive-Sentiment FPR gaps for the Amazon dataset are shown in Figure \ref{amazon_gaps_fig}. For most models, the Negative-Sentiment FPR gap for White-associated text is significantly above zero at 95\% confidence. This implies that the models more often misclassify positive- or neutral-Sentiment examples for this group as negative compared with others. For large OPT, Llama-2 and Mistral LLMs, a similar but smaller elevation in this gap is observed for examples associated with African Americans and Asians. For the Positive-Sentiment FPR gap, a significant negative value is observed for all models. More recent models, Llama-3 and Mistral, exhibit some of the largest negative gaps. Combined with an elevated Negative-Sentiment FPR gap, this implies that the models tend to view examples from the White group in a negative light more often than other groups.

\begin{figure}[ht!]
\centering
\includegraphics[width=0.99\textwidth]{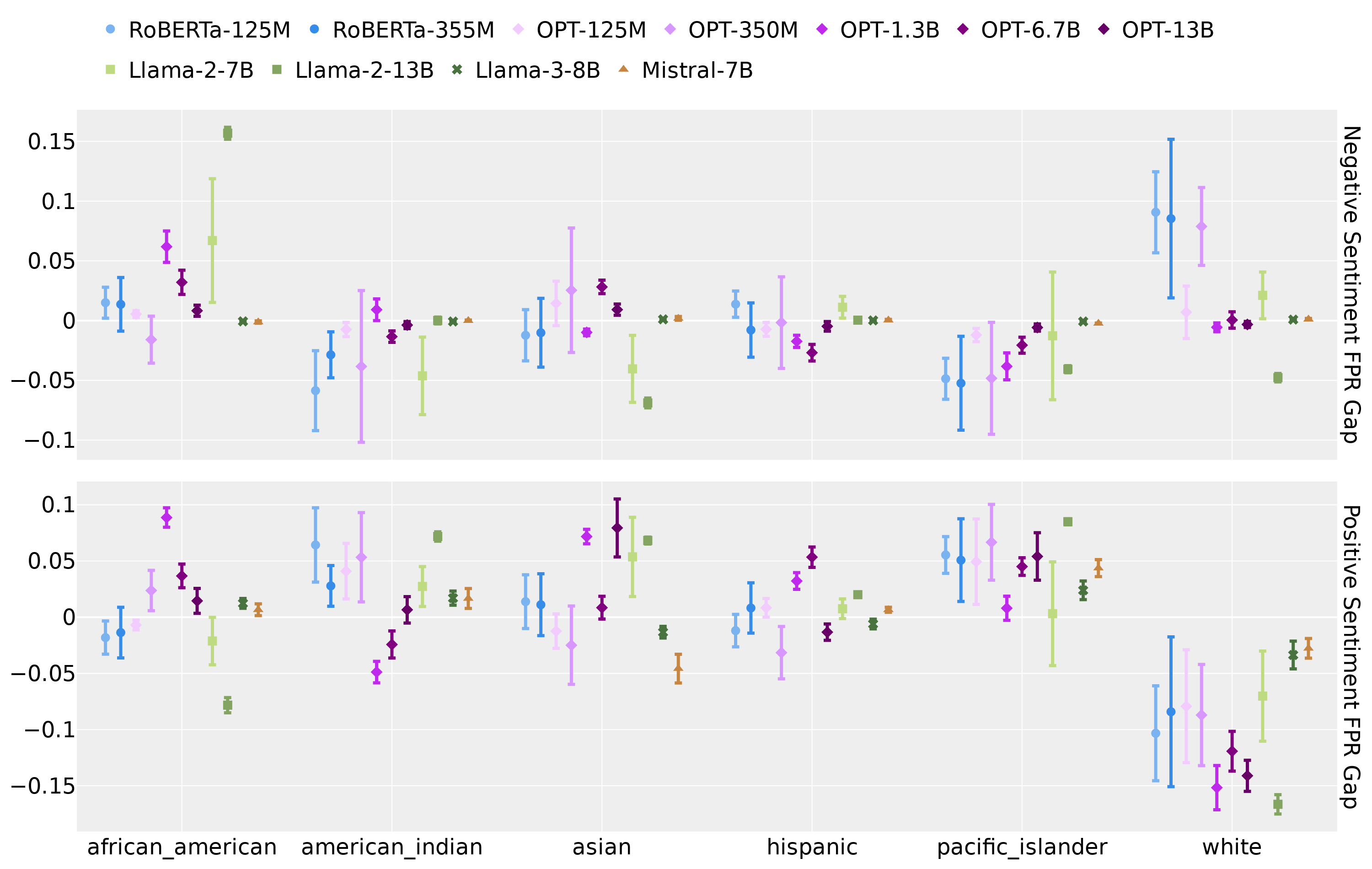}
\caption{Negative- and Positive-Sentiment FPR gaps as measured by the NS-Prompts dataset.}
\label{ns_prompt_gaps_fig} 
\end{figure}

Figure \ref{ns_prompt_gaps_fig} displays the measured gaps for the NS-Prompts dataset. Recall that all labels for this dataset are neutral. Thus, any non-neutral predictions are, by construction, incorrect. When considering RoBERTa and Llama-2 models, the identified gaps share similarities with the African-American group. That is, elevated Negative-Sentiment FPR gaps and Positive-Sentiment FPR gaps below zero. While the Negative-Sentiment FPR gaps for other models are near zero for White examples, all models produce negative and statistically significant Positive-Sentiment FPR gaps. This implies that neutral examples associated with White race are construed as positive at much lower rates relative to other groups.

\begin{figure}[ht!]
\centering
\includegraphics[width=0.99\textwidth]{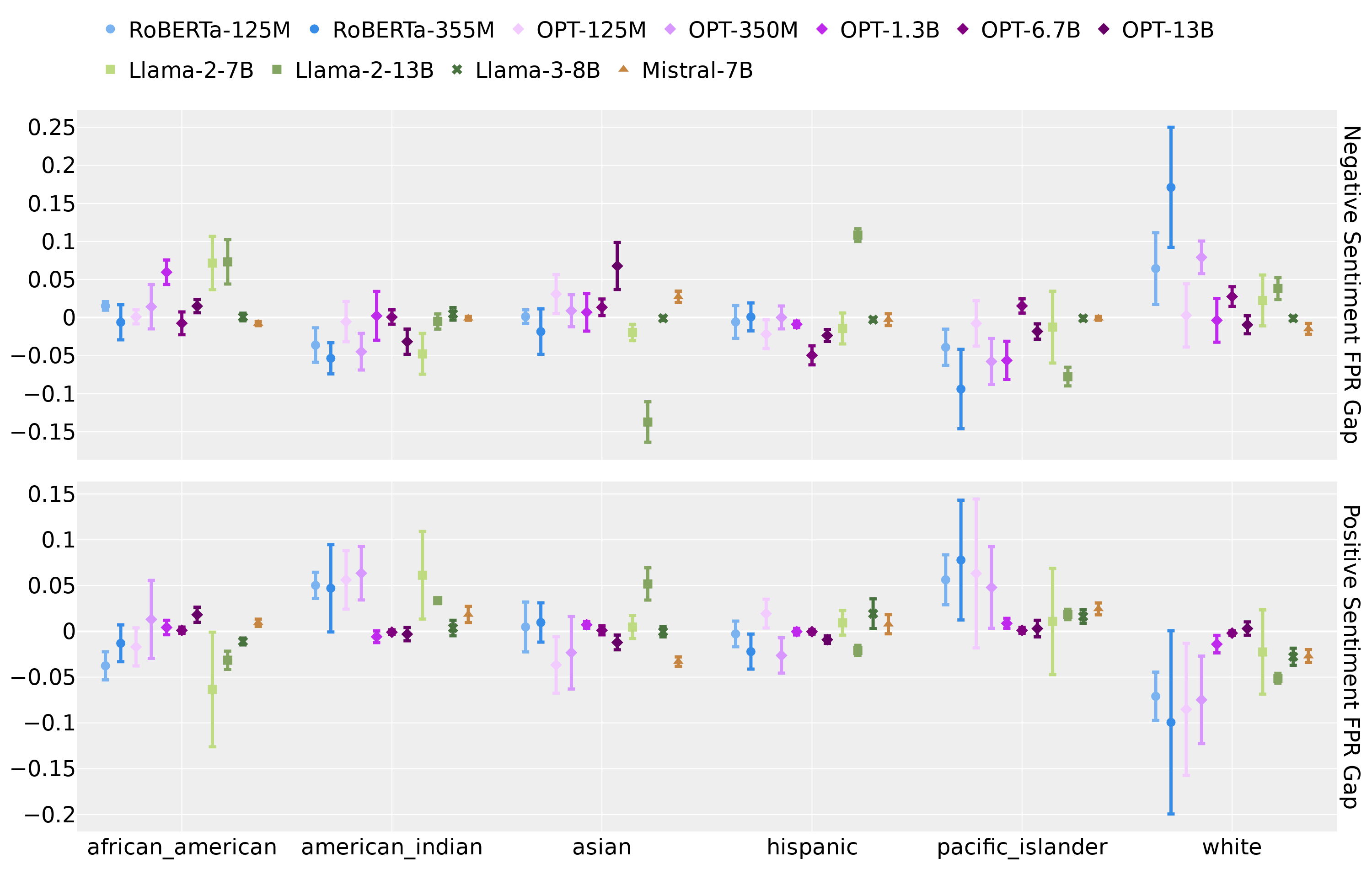}
\caption{Negative- and Positive-Sentiment FPR gaps as measured by the Regard dataset.}
\label{regard_gaps_fig} 
\end{figure}

Results for the Regard dataset reveal similar trends to the Amazon and NS-Prompts experiments. However, the gaps, displayed in Figure \ref{regard_gaps_fig}, are somewhat smaller. As in previous measurements, White-associated texts see elevated Negative-Sentiment FPR gaps and Positive-Sentiment FPR gaps below zero for many models. Furthermore, strong parallels exist for the gaps observed for text associated with African Americans. This is especially true for RoBERTa, small OPT, Llama-2, and Llama-3 models, where the gaps for these groups are highly correlated.

The measurements in these results are surprising. However, as discussed in detail in Section \ref{discussion} below, the gaps observed for the White group are not believed to be reflections of true bias. Rather, we conjecture that they are an artifact of a mismatch between the template-based probing datasets that explicitly reference race in the text of samples to link membership and the presence of markedness in LLM pretraining data.

\subsection{Prompt-Based Results}

The results in Section \ref{fine_tuning_results_discussion} exhibit clear anomalies when measuring performance gaps using template-based probes. A natural question is whether such irregularities arise due to the task-specific fine-tuning step or represent an intrinsic quality of the LLMs. To further isolate the issue to LLM pretraining, prompting is used to perform sentiment classification for the Amazon dataset, shedding the need for fine-tuning. The experiments are limited to decoder-only models of sufficient size to ensure that classification performance adequately exceeds that of a random classifier. 

The average classification accuracy of the prompting and fine-tuning approaches on the Amazon dataset is reported in Appendix \ref{appendix_prompt_finetune_tables}. Generally, the accuracy of prompt-based classification is lower than the fine-tuning counterpart. This is especially true for the oldest model, OPT. However, newer models still provide good performance through prompting. Notably, Qwen-2.5 produces very strong classification accuracy with a 9-shot prompt drawn from SemEval at $92.3\%$. Other models also perform fairly well using few-shot prompting. Regardless, as classifiers, all prompted LLMs perform well above a random model. Perhaps due to model size or limited reasoning tokens, zero-shot CoT does not significantly improve performance \citep{Wei1}.

\begin{figure}[ht!]
\centering
\includegraphics[width=0.99\textwidth]{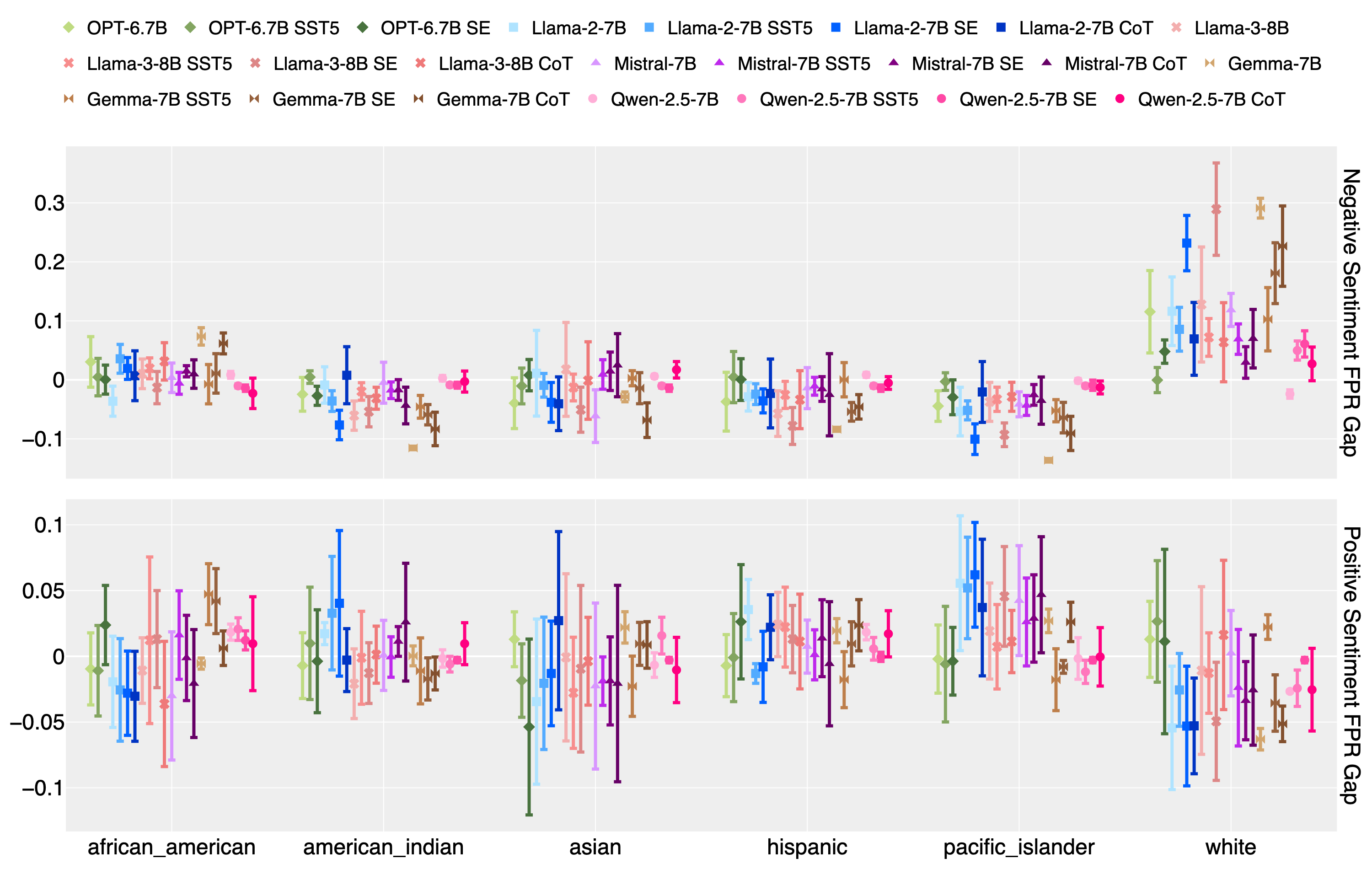}
\caption{Negative- and Positive-Sentiment FPR gaps as measured by the Amazon dataset with prompt-based classification. In the legend, model names without a suffix indicate zero-shot prompting. SST5 and SE indicate 9-shot prompts with examples drawn from the SST5 and SemEval datasets, respectively.}
\label{prompt_gaps_fig} 
\end{figure}

As in Section \ref{fine_tuning_results_discussion}, Negative- and Positive-Sentiment FPR gaps are computed for each LLM's predictions. These gaps are exhibited in Figure \ref{prompt_gaps_fig}. Due to the lower accuracy and generation volatility, the gap CIs are visibly wider than those of the fine-tuning experiments. Nonetheless, a clear and familiar pattern is seen in these results. Positive mean gaps in Negative-Sentiment FPR are present across a majority of examples for African American and White races. Similarly, negative mean gaps for Positive-Sentiment FPR are measured for both races in most settings. The consistency between these results and those of the fine-tuning experiments strongly suggests that the irregularities present in the template-based measurements are not the result of the fine-tuning stage, but are, rather, an expression of an intrinsic aspect of the LLMs themselves.

\section{Discussion} \label{discussion}

Across the experiments an overall tendency of the models to classify White-associated text as exhibiting more negative sentiment at a higher rate than other groups is observed. The trends in the results above are consistent between model type, model size, template-based probing dataset, and even classification strategy. The overall agreement of the prompting and fine-tuning results indicates that the observed gaps are not linked to idiosyncrasies in the fine-tuning process but are, rather, more fundamental to the LLMs themselves and the design of the template-based probes. In addition, the models chosen for experimentation are base or instruction-tuned versions. That is, their predictions are not influenced by interceding alignment techniques \citep{bai2022training, rafailov2023direct}, which might otherwise obscure behavior learned during pretraining. Rather than implying an extant bias, we hypothesize below that this phenomenon is due to an interaction between the structure of the templates used in the measurement of bias and markedness in LLM pretraining data. Regardless of the underlying cause, these observations should lead us to re-think the clarity of counterfactual bias analysis in this context.

\subsection{Markedness and Template-Based Probes} \label{markedness_conjecture}

English pretraining data for LLMs is dominated by text drawn from areas where the racial majority is White \citep{Bender1, Navigli1}. Several studies have confirmed that markedness is widespread in internet data, with White race and male gender constituting the unmarked defaults \citep{Wolfe1, Bailey1}. Furthermore, it has been shown that models, and LLMs in particular, trained on web data reflect these markedness characteristics \citep{Bender1, Wolfe2, Wolfe1, cheng-etal-2023-marked}. On the other hand, in templates commonly used for bias quantification, race is explicitly mentioned to establish group membership. As such, template-based text that explicitly establishes that the subject is ``White'' essentially constitute out-of-domain examples \citep{blodgett2021stereotyping, dressler1985predictiveness}. Such a mismatch likely influences model predictions.

We hypothesize that the disparities observed in Section \ref{results} associated with the White group are due to the prevalence of markedness in LLM pretraining text. A key assumption underlying unmarked representations is that humans are adept at recognizing unstated implications in text. LLMs trained solely on unstructured next-token prediction, which underpins almost all modern LLM pretraining, may lack the ability to perceive such implications, resulting in surprising behavior. Using templates that represent group membership through explicit description likely makes certain text appear uncommon for traditionally unmarked groups. As such, these templates may lead to artificially elevated error rates in LLMs, skewing bias measurements in unpredictable ways and clouding the lens provided by datasets of this structure. To this end, Appendix \ref{sexuality_and_gender_results} provides an extended set of results showing that similar irregularities arise when considering unmarked groups for sexuality and gender, providing additional supporting evidence for this hypothesis.

Including datasets that explicitly correct for markedness in LLM pretraining could better align template-based text. Appendix \ref{appendix_span_table} suggests that more recent LLMs, trained on larger multilingual datasets, show improvements in measured gap sizes. Both Llama-3-8B and Mistral-7B have the smallest difference between the most positive and negative gaps for Negative-Sentiment FPR, averaged over the three datasets. Llama-3-8B also produces the lowest average difference for Positive-Sentiment FPR. Given that White-group gaps often rank among the extremes, this suggests newer models may be less affected by markedness.

The studies and results above, and in the appendix, suggest that markedness may indeed play a role in the experimental observations of this work. However, empirical validation of the conjecture that markedness is the mechanism introducing the effects observed in Section \ref{results} likely requires that an LLM be trained from scratch using a dataset with ``consistent'' marking. Thereafter, the gaps across demographics would be re-assessed using the same template-based probes. The size and availability of LLM pretraining data make constructing such a dataset quite challenging. Moreover, the computational resources required to properly pretrain an LLM are substantial. As such, this valuable study is deferred to future work.

\section{Conclusions and Future Work}

This paper presents unexpected, and likely flawed, bias measurements related to race when using template-based bias probes. The measurements remain consistent across a number of different experimental settings and varied datasets. Rather than indicating genuine social bias in the LLMs, we conjecture that these outliers stem from a misalignment between template-based bias probes and LLM pretraining data due to markedness. Regardless of the underlying cause, these findings highlight the need to consider the impact that the use of bias probes relying on marked text has on the measurement of bias. In this case, such probes produce largely misleading results. 

Assuming that linguistic markedness contributes significantly to the measurements in this work, which requires further investigation to confirm, several avenues for mitigating such effects when using template-based probes are worth exploring. LLMs trained on a more global representation of text, where majority demographics differ and marked groups vary, could improve the robustness of such models when encountering explicit demographic descriptors. Another approach is to strive to control for the impact of markedness by designing evaluation setups that test both unmarked and explicitly marked versions of the same text (for example, comparing ``a CEO'' and ``a White CEO'') or by using neutral placeholders like [RACE] to isolate the impact of demographic terms and potentially correct for the flaws identified in this work. There are, however, challenges with this approach. One needs to identify which groups are considered unmarked from the ``perspective'' of the LLM, requiring detailed knowledge of the underlying pretraining data. In addition, the unassociated text is unlikely to solely represent the unmarked group, but rather a mix of representations. Ideally, artificial injection of demographic information would not be required. For example, the studies of \citet{seyyed2020chexclusion} and \citet{sap-etal-2019-risk} establish group membership through meta-data, self-identification, or classification techniques rather than explicitly in text. These methods avoid the out-of-domain nature of template-based examples of the kind studied here and do not see the unnatural patterns we observed.  Future work will design experiments to validate the misalignment due to markedness conjecture and construct straightforward ways to mitigate such issues in LLMs.

\bibliography{main}

@incollection{de2019bias,
  address = {USA},
  title={Bias in bios: a case study of semantic representation bias in a high-stakes setting},
  author={De-Arteaga, Maria and Romanov, Alexey and Wallach, Hanna and Chayes, Jennifer and Borgs, Christian and Chouldechova, Alexandra and Geyik, Sahin and Kenthapadi, Krishnaram and Kalai, Adam Tauman},
  booktitle ={Proceedings of the Conference on Fairness, Accountability, and Transparency},
  note = {Atlanta, GA},
  series = {{FAT*}'19},
  pages = {120--128},
  publisher = "Association for Computing Machinery",
  year={2019},
}

@misc{touvron2023llama,
      title={Llama 2: Open Foundation and Fine-Tuned Chat Models}, 
      author={Hugo Touvron and Louis Martin and Kevin Stone and Peter Albert and Amjad Almahairi and Yasmine Babaei and Nikolay Bashlykov and Soumya Batra and Prajjwal Bhargava and Shruti Bhosale and Dan Bikel and Lukas Blecher and Cristian Canton Ferrer and Moya Chen and Guillem Cucurull and David Esiobu and Jude Fernandes and Jeremy Fu and Wenyin Fu and Brian Fuller and Cynthia Gao and Vedanuj Goswami and Naman Goyal and Anthony Hartshorn and Saghar Hosseini and Rui Hou and Hakan Inan and Marcin Kardas and Viktor Kerkez and Madian Khabsa and Isabel Kloumann and Artem Korenev and Punit Singh Koura and Marie-Anne Lachaux and Thibaut Lavril and Jenya Lee and Diana Liskovich and Yinghai Lu and Yuning Mao and Xavier Martinet and Todor Mihaylov and Pushkar Mishra and Igor Molybog and Yixin Nie and Andrew Poulton and Jeremy Reizenstein and Rashi Rungta and Kalyan Saladi and Alan Schelten and Ruan Silva and Eric Michael Smith and Ranjan Subramanian and Xiaoqing Ellen Tan and Binh Tang and Ross Taylor and Adina Williams and Jian Xiang Kuan and Puxin Xu and Zheng Yan and Iliyan Zarov and Yuchen Zhang and Angela Fan and Melanie Kambadur and Sharan Narang and Aurelien Rodriguez and Robert Stojnic and Sergey Edunov and Thomas Scialom},
      year={2023},
note = "Preprint at \url{https://arxiv.org/abs/2307.09288}"
}

@incollection{shwartz2020neural,
    title = "Do Neural Language Models Overcome Reporting Bias?",
    author = "Shwartz, Vered  and
      Choi, Yejin",
    editor = "Scott, Donia  and
      Bel, Nuria  and
      Zong, Chengqing",
    booktitle = "Proceedings of the 28th International Conference on Computational Linguistics",
    month = dec,
    year = "2020",
    address = "Barcelona, Spain (Online)",
    publisher = "International Committee on Computational Linguistics",
    url = "https://aclanthology.org/2020.coling-main.605",
    doi = "10.18653/v1/2020.coling-main.605",
    pages = "6863--6870",
}

@incollection{
hu2022lora,
title={Lo{RA}: Low-Rank Adaptation of Large Language Models},
author={Edward J Hu and Yelong Shen and Phillip Wallis and Zeyuan Allen-Zhu and Yuanzhi Li and Shean Wang and Lu Wang and Weizhu Chen},
booktitle={International Conference on Learning Representations},
year={2022},
publisher={International Conference on Learning Representations},
}

@incollection{sheng2019woman,
    title = "The Woman Worked as a Babysitter: On Biases in Language Generation",
    author = "Sheng, Emily  and
      Chang, Kai-Wei  and
      Natarajan, Premkumar  and
      Peng, Nanyun",
    editor = "Inui, Kentaro  and
      Jiang, Jing  and
      Ng, Vincent  and
      Wan, Xiaojun",
    booktitle = "Proceedings of the 2019 Conference on Empirical Methods in Natural Language Processing and the 9th International Joint Conference on Natural Language Processing (EMNLP-IJCNLP)",
    month = nov,
    year = "2019",
    address = "Hong Kong, China",
    publisher = "Association for Computational Linguistics",
    url = "https://aclanthology.org/D19-1339",
    doi = "10.18653/v1/D19-1339",
    pages = "3407--3412",
}

@incollection{alnegheimish2022using,
    title = "Using Natural Sentence Prompts for Understanding Biases in Language Models",
    author = "Alnegheimish, Sarah  and
      Guo, Alicia  and
      Sun, Yi",
    editor = "Carpuat, Marine  and
      de Marneffe, Marie-Catherine  and
      Meza Ruiz, Ivan Vladimir",
    booktitle = "Proceedings of the 2022 Conference of the North American Chapter of the Association for Computational Linguistics: Human Language Technologies",
    month = jul,
    year = "2022",
    address = "Seattle, United States",
    publisher = "Association for Computational Linguistics",
    url = "https://aclanthology.org/2022.naacl-main.203",
    doi = "10.18653/v1/2022.naacl-main.203",
    pages = "2824--2830",
}

@article{czarnowska2021quantifying,
  title={Quantifying social biases in {NLP}: A generalization and empirical comparison of extrinsic fairness metrics},
  author={Czarnowska, Paula and Vyas, Yogarshi and Shah, Kashif},
  journal={Transactions of the Association for Computational Linguistics},
  volume={9},
  pages={1249--1267},
  year={2021},
  publisher={MIT Press One Rogers Street, Cambridge, MA 02142-1209, USA journals-info~…}
}

@misc{Rae1,
      title={Scaling Language Models: Methods, Analysis \& Insights from Training Gopher}, 
      author={Jack W. Rae and Sebastian Borgeaud and Trevor Cai and Katie Millican and Jordan Hoffmann and Francis Song and John Aslanides and Sarah Henderson and Roman Ring and Susannah Young and Eliza Rutherford and Tom Hennigan and Jacob Menick and Albin Cassirer and Richard Powell and George van den Driessche and Lisa Anne Hendricks and Maribeth Rauh and Po-Sen Huang and Amelia Glaese and Johannes Welbl and Sumanth Dathathri and Saffron Huang and Jonathan Uesato and John Mellor and Irina Higgins and Antonia Creswell and Nat McAleese and Amy Wu and Erich Elsen and Siddhant Jayakumar and Elena Buchatskaya and David Budden and Esme Sutherland and Karen Simonyan and Michela Paganini and Laurent Sifre and Lena Martens and Xiang Lorraine Li and Adhiguna Kuncoro and Aida Nematzadeh and Elena Gribovskaya and Domenic Donato and Angeliki Lazaridou and Arthur Mensch and Jean-Baptiste Lespiau and Maria Tsimpoukelli and Nikolai Grigorev and Doug Fritz and Thibault Sottiaux and Mantas Pajarskas and Toby Pohlen and Zhitao Gong and Daniel Toyama and Cyprien de Masson d'Autume and Yujia Li and Tayfun Terzi and Vladimir Mikulik and Igor Babuschkin and Aidan Clark and Diego de Las Casas and Aurelia Guy and Chris Jones and James Bradbury and Matthew Johnson and Blake Hechtman and Laura Weidinger and Iason Gabriel and William Isaac and Ed Lockhart and Simon Osindero and Laura Rimell and Chris Dyer and Oriol Vinyals and Kareem Ayoub and Jeff Stanway and Lorrayne Bennett and Demis Hassabis and Koray Kavukcuoglu and Geoffrey Irving},
      year={2022},
      eprint={2112.11446},
      archivePrefix={arXiv},
      primaryClass={cs.CL},
      url={https://arxiv.org/abs/2112.11446}, 
}

@misc{ganguli2023capacity,
      title={The Capacity for Moral Self-Correction in Large Language Models}, 
      author={Deep Ganguli and Amanda Askell and Nicholas Schiefer and Thomas I. Liao and Kamilė Lukošiūtė and Anna Chen and Anna Goldie and Azalia Mirhoseini and Catherine Olsson and Danny Hernandez and Dawn Drain and Dustin Li and Eli Tran-Johnson and Ethan Perez and Jackson Kernion and Jamie Kerr and Jared Mueller and Joshua Landau and Kamal Ndousse and Karina Nguyen and Liane Lovitt and Michael Sellitto and Nelson Elhage and Noemi Mercado and Nova DasSarma and Oliver Rausch and Robert Lasenby and Robin Larson and Sam Ringer and Sandipan Kundu and Saurav Kadavath and Scott Johnston and Shauna Kravec and Sheer El Showk and Tamera Lanham and Timothy Telleen-Lawton and Tom Henighan and Tristan Hume and Yuntao Bai and Zac Hatfield-Dodds and Ben Mann and Dario Amodei and Nicholas Joseph and Sam McCandlish and Tom Brown and Christopher Olah and Jack Clark and Samuel R. Bowman and Jared Kaplan},
      year={2023},
      eprint={2302.07459},
      archivePrefix={arXiv},
      primaryClass={cs.CL},
      url={https://arxiv.org/abs/2302.07459}, 
}

@misc{bai2022training,
      title={Training a Helpful and Harmless Assistant with Reinforcement Learning from Human Feedback}, 
      author={Yuntao Bai and Andy Jones and Kamal Ndousse and Amanda Askell and Anna Chen and Nova DasSarma and Dawn Drain and Stanislav Fort and Deep Ganguli and Tom Henighan and Nicholas Joseph and Saurav Kadavath and Jackson Kernion and Tom Conerly and Sheer El-Showk and Nelson Elhage and Zac Hatfield-Dodds and Danny Hernandez and Tristan Hume and Scott Johnston and Shauna Kravec and Liane Lovitt and Neel Nanda and Catherine Olsson and Dario Amodei and Tom Brown and Jack Clark and Sam McCandlish and Chris Olah and Ben Mann and Jared Kaplan},
      year={2022},
  note = "Preprint at \url{https://arxiv.org/abs/2204.05862}"
}

@misc{
liu2020roberta,
title={Ro{BERT}a: A Robustly Optimized {BERT} Pretraining Approach},
author={Yinhan Liu and Myle Ott and Naman Goyal and Jingfei Du and Mandar Joshi and Danqi Chen and Omer Levy and Mike Lewis and Luke Zettlemoyer and Veselin Stoyanov},
year={2020},
url={https://openreview.net/forum?id=SyxS0T4tvS}
}

@article{Navigli1,
author = {Navigli, Roberto and Conia, Simone and Ross, Bj\"{o}rn},
title = {Biases in Large Language Models: Origins, Inventory, and Discussion},
year = {2023},
issue_date = {June 2023},
publisher = {Association for Computing Machinery},
address = {New York, NY, USA},
volume = {15},
number = {2},
issn = {1936-1955},
url = {https://doi.org/10.1145/3597307},
doi = {10.1145/3597307},
journal = {J. Data and Information Quality},
month = {jun},
articleno = {10},
numpages = {21}
}

@incollection{Bender1,
author = {Bender, Emily M. and Gebru, Timnit and McMillan-Major, Angelina and Shmitchell, Shmargaret},
title = {On the Dangers of Stochastic Parrots: Can Language Models Be Too Big?},
year = {2021},
isbn = {9781450383097},
publisher = {Association for Computing Machinery},
address = {New York, NY, USA},
url = {https://doi.org/10.1145/3442188.3445922},
doi = {10.1145/3442188.3445922},
booktitle = {Proceedings of the 2021 ACM Conference on Fairness, Accountability, and Transparency},
pages = {610–623},
numpages = {14},
location = {Virtual Event, Canada},
series = {FAccT '21}
}

@incollection{sap-etal-2019-risk,
    title = "The Risk of Racial Bias in Hate Speech Detection",
    author = "Sap, Maarten  and
      Card, Dallas  and
      Gabriel, Saadia  and
      Choi, Yejin  and
      Smith, Noah A.",
    editor = "Korhonen, Anna  and
      Traum, David  and
      M{\`a}rquez, Llu{\'\i}s",
    booktitle = "Proceedings of the 57th Annual Meeting of the Association for Computational Linguistics",
    month = jul,
    year = "2019",
    address = "Florence, Italy",
    publisher = "Association for Computational Linguistics",
    url = "https://aclanthology.org/P19-1163",
    doi = "10.18653/v1/P19-1163",
    pages = "1668--1678",
}

@incollection{seyyed2020chexclusion,
  title={CheXclusion: Fairness gaps in deep chest X-ray classifiers},
  author={Seyyed-Kalantari, Laleh and Liu, Guanxiong and McDermott, Matthew and Chen, Irene Y and Ghassemi, Marzyeh},
  booktitle={BIOCOMPUTING 2021: proceedings of the Pacific symposium},
  pages={232--243},
  year={2020},
  publisher={World Scientific Publishing Company}
}

@incollection{wan2023kelly,
    title = "``{Kelly} is a Warm Person, {J}oseph is a Role Model'': Gender Biases in {LLM}-Generated Reference Letters",
    author = "Wan, Yixin  and
      Pu, George  and
      Sun, Jiao  and
      Garimella, Aparna  and
      Chang, Kai-Wei  and
      Peng, Nanyun",
    editor = "Bouamor, Houda  and
      Pino, Juan  and
      Bali, Kalika",
    booktitle = "Findings of the Association for Computational Linguistics: EMNLP 2023",
    month = dec,
    year = "2023",
    address = "Singapore",
    publisher = "Association for Computational Linguistics",
    url = "https://aclanthology.org/2023.findings-emnlp.243",
    doi = "10.18653/v1/2023.findings-emnlp.243",
    pages = "3730--3748",
}

@article{10.1162/coli_a_00524,
    author = {Gallegos, Isabel O. and Rossi, Ryan A. and Barrow, Joe and Tanjim, Md Mehrab and Kim, Sungchul and Dernoncourt, Franck and Yu, Tong and Zhang, Ruiyi and Ahmed, Nesreen K.},
    title = {Bias and Fairness in Large Language Models: A Survey},
    journal = {Computational Linguistics},
    volume = {50},
    number = {3},
    pages = {1097-1179},
    year = {2024},
    month = {09},
    issn = {0891-2017},
    doi = {10.1162/coli_a_00524},
    url = {https://doi.org/10.1162/coli\_a\_00524},
    eprint = {https://direct.mit.edu/coli/article-pdf/50/3/1097/2471010/coli\_a\_00524.pdf},
}

@misc{zhang2022opt,
      title={{OPT}: Open Pre-trained Transformer Language Models}, 
      author={Susan Zhang and Stephen Roller and Naman Goyal and Mikel Artetxe and Moya Chen and Shuohui Chen and Christopher Dewan and Mona Diab and Xian Li and Xi Victoria Lin and Todor Mihaylov and Myle Ott and Sam Shleifer and Kurt Shuster and Daniel Simig and Punit Singh Koura and Anjali Sridhar and Tianlu Wang and Luke Zettlemoyer},
      year={2022},
note = "Preprint at \url{https://arxiv.org/abs/2205.01068}"
}

@misc{jiang2023mistral7b,
      title={Mistral 7B}, 
      author={Albert Q. Jiang and Alexandre Sablayrolles and Arthur Mensch and Chris Bamford and Devendra Singh Chaplot and Diego de las Casas and Florian Bressand and Gianna Lengyel and Guillaume Lample and Lucile Saulnier and Lélio Renard Lavaud and Marie-Anne Lachaux and Pierre Stock and Teven Le Scao and Thibaut Lavril and Thomas Wang and Timothée Lacroix and William El Sayed},
      year={2023},
note = "Preprint at \url{https://arxiv.org/abs/2310.06825}"
}

@incollection{delobelle2022measuring,
    title = "Measuring Fairness with Biased Rulers: A Comparative Study on Bias Metrics for Pre-trained Language Models",
    author = "Delobelle, Pieter  and
      Tokpo, Ewoenam  and
      Calders, Toon  and
      Berendt, Bettina",
    editor = "Carpuat, Marine  and
      de Marneffe, Marie-Catherine  and
      Meza Ruiz, Ivan Vladimir",
    booktitle = "Proceedings of the 2022 Conference of the North American Chapter of the Association for Computational Linguistics: Human Language Technologies",
    month = jul,
    year = "2022",
    address = "Seattle, United States",
    publisher = "Association for Computational Linguistics",
    url = "https://aclanthology.org/2022.naacl-main.122",
    doi = "10.18653/v1/2022.naacl-main.122",
    pages = "1693--1706",
}

@article{mokander2023auditing,
  title={Auditing large language models: A three-layered approach},
  author={Jakob M{\"o}kander and Jonas Schuett and Hannah Rose Kirk and Luciano Floridi},
  journal={AI and Ethics},
  pages={1--31},
  year={2023},
  publisher={Springer}
}

@incollection{liang2021towards,
  title={Towards Understanding and Mitigating Social Biases in Language Models},
  author={Liang, Paul Pu and Wu, Chiyu and Morency, Louis-Philippe and Salakhutdinov, Ruslan},
  booktitle={Proceedings of the 38th International Conference on Machine Learning},
  pages={6565--6576},
  year={2021},
  publisher="PMLR",
}

@incollection{sheng2021societal,
    title = "Societal Biases in Language Generation: Progress and Challenges",
    author = "Sheng, Emily  and
      Chang, Kai-Wei  and
      Natarajan, Prem  and
      Peng, Nanyun",
    editor = "Zong, Chengqing  and
      Xia, Fei  and
      Li, Wenjie  and
      Navigli, Roberto",
    booktitle = "Proceedings of the 59th Annual Meeting of the Association for Computational Linguistics and the 11th International Joint Conference on Natural Language Processing (Volume 1: Long Papers)",
    month = aug,
    year = "2021",
    address = "Online",
    publisher = "Association for Computational Linguistics",
    url = "https://aclanthology.org/2021.acl-long.330",
    doi = "10.18653/v1/2021.acl-long.330",
    pages = "4275--4293",
}

@misc{liu2023trustworthy,
  title={Trustworthy {LLMs}: a Survey and Guideline for Evaluating Large Language Models' Alignment},
  author={Liu, Yang and Yao, Yuanshun and Ton, Jean-Francois and Zhang, Xiaoying and Cheng, Ruocheng Guo Hao and Klochkov, Yegor and Taufiq, Muhammad Faaiz and Li, Hang},
  year={2023},
note = "Preprint at \url{https://arxiv.org/abs/2308.05374}"
}

@misc{kaneko2024evaluating,
  title={Evaluating Gender Bias in Large Language Models via Chain-of-Thought Prompting},
  author={Kaneko, Masahiro and Bollegala, Danushka and Okazaki, Naoaki and Baldwin, Timothy},
  year={2024},
note = "Preprint at \url{https://arxiv.org/abs/2401.15585}"
}

@misc{echterhoff2024cognitive,
  title={Cognitive Bias in High-Stakes Decision-Making with {LLMs}},
  author={Echterhoff, Jessica and Liu, Yao and Alessa, Abeer and McAuley, Julian and He, Zexue},
  year={2024},
note = "Preprint at \url{https://arxiv.org/abs/2403.00811}"
}

@incollection{Wei1,
author = {Wei, Jason and Wang, Xuezhi and Schuurmans, Dale and Bosma, Maarten and Ichter, Brian and Xia, Fei and Chi, Ed H. and Le, Quoc V. and Zhou, Denny},
title = {Chain-of-thought prompting elicits reasoning in large language models},
year = {2024},
isbn = {9781713871088},
publisher = {Curran Associates Inc.},
address = {Red Hook, NY, USA},
booktitle = {Proceedings of the 36th International Conference on Neural Information Processing Systems},
articleno = {1800},
numpages = {14},
location = {New Orleans, LA, USA},
series = {NIPS '22}
}

@incollection{Kojima1,
author = {Kojima, Takeshi and Gu, Shixiang Shane and Reid, Machel and Matsuo, Yutaka and Iwasawa, Yusuke},
title = {Large language models are zero-shot reasoners},
year = {2024},
isbn = {9781713871088},
publisher = {Curran Associates Inc.},
address = {Red Hook, NY, USA},
booktitle = {Proceedings of the 36th International Conference on Neural Information Processing Systems},
articleno = {1613},
numpages = {15},
location = {New Orleans, LA, USA},
series = {NIPS '22}
}

@article{
liang2023holistic,
title={Holistic Evaluation of Language Models},
author={Percy Liang and Rishi Bommasani and Tony Lee and Dimitris Tsipras and Dilara Soylu and Michihiro Yasunaga and Yian Zhang and Deepak Narayanan and Yuhuai Wu and Ananya Kumar and Benjamin Newman and Binhang Yuan and Bobby Yan and Ce Zhang and Christian Alexander Cosgrove and Christopher D Manning and Christopher Re and Diana Acosta-Navas and Drew Arad Hudson and Eric Zelikman and Esin Durmus and Faisal Ladhak and Frieda Rong and Hongyu Ren and Huaxiu Yao and Jue WANG and Keshav Santhanam and Laurel Orr and Lucia Zheng and Mert Yuksekgonul and Mirac Suzgun and Nathan Kim and Neel Guha and Niladri S. Chatterji and Omar Khattab and Peter Henderson and Qian Huang and Ryan Andrew Chi and Sang Michael Xie and Shibani Santurkar and Surya Ganguli and Tatsunori Hashimoto and Thomas Icard and Tianyi Zhang and Vishrav Chaudhary and William Wang and Xuechen Li and Yifan Mai and Yuhui Zhang and Yuta Koreeda},
journal={Transactions on Machine Learning Research},
issn={2835-8856},
year={2023},
url={https://openreview.net/forum?id=iO4LZibEqW},
note={Featured Certification, Expert Certification}
}

@incollection{socher2013recursive,
    title = "Recursive Deep Models for Semantic Compositionality Over a Sentiment Treebank",
    author = "Socher, Richard  and
      Perelygin, Alex  and
      Wu, Jean  and
      Chuang, Jason  and
      Manning, Christopher D.  and
      Ng, Andrew  and
      Potts, Christopher",
    editor = "Yarowsky, David  and
      Baldwin, Timothy  and
      Korhonen, Anna  and
      Livescu, Karen  and
      Bethard, Steven",
    booktitle = "Proceedings of the 2013 Conference on Empirical Methods in Natural Language Processing",
    month = oct,
    year = "2013",
    address = "Seattle, Washington, USA",
    publisher = "Association for Computational Linguistics",
    url = "https://aclanthology.org/D13-1170",
    pages = "1631--1642",
}

@misc{tian2023interpretable,
  title={Interpretable stereotype identification through reasoning},
  author={Tian, Jacob-Junqi and Dige, Omkar and Emerson, David and Khattak, Faiza Khan},
  year={2023},
note = "Preprint at \url{https://arxiv.org/abs/2308.00071}"
}

@incollection{cheng-etal-2023-marked,
    title = "Marked Personas: Using Natural Language Prompts to Measure Stereotypes in Language Models",
    author = "Cheng, Myra  and
      Durmus, Esin  and
      Jurafsky, Dan",
    editor = "Rogers, Anna  and
      Boyd-Graber, Jordan  and
      Okazaki, Naoaki",
    booktitle = "Proceedings of the 61st Annual Meeting of the Association for Computational Linguistics (Volume 1: Long Papers)",
    month = jul,
    year = "2023",
    address = "Toronto, Canada",
    publisher = "Association for Computational Linguistics",
    url = "https://aclanthology.org/2023.acl-long.84",
    doi = "10.18653/v1/2023.acl-long.84",
    pages = "1504--1532",
}

@incollection{blodgett2021stereotyping,
    title = "Stereotyping {N}orwegian Salmon: An Inventory of Pitfalls in Fairness Benchmark Datasets",
    author = "Blodgett, Su Lin  and
      Lopez, Gilsinia  and
      Olteanu, Alexandra  and
      Sim, Robert  and
      Wallach, Hanna",
    editor = "Zong, Chengqing  and
      Xia, Fei  and
      Li, Wenjie  and
      Navigli, Roberto",
    booktitle = "Proceedings of the 59th Annual Meeting of the Association for Computational Linguistics and the 11th International Joint Conference on Natural Language Processing (Volume 1: Long Papers)",
    month = aug,
    year = "2021",
    address = "Online",
    publisher = "Association for Computational Linguistics",
    url = "https://aclanthology.org/2021.acl-long.81",
    doi = "10.18653/v1/2021.acl-long.81",
    pages = "1004--1015",
}

@article{waugh1982marked,
  title={Marked and unmarked: A choice between unequals in semiotic structure},
  author={Waugh, Linda R},
  year={1982},
  journal={Linguistics},
  publisher={Walter de Gruyter, Berlin/New York Berlin, New York}
}

@incollection{SemEval2018Task1,
    title = "{S}em{E}val-2018 Task 1: Affect in Tweets",
    author = "Mohammad, Saif  and
      Bravo-Marquez, Felipe  and
      Salameh, Mohammad  and
      Kiritchenko, Svetlana",
    editor = "Apidianaki, Marianna  and
      Mohammad, Saif M.  and
      May, Jonathan  and
      Shutova, Ekaterina  and
      Bethard, Steven  and
      Carpuat, Marine",
    booktitle = "Proceedings of the 12th International Workshop on Semantic Evaluation",
    month = jun,
    year = "2018",
    address = "New Orleans, Louisiana",
    publisher = "Association for Computational Linguistics",
    url = "https://aclanthology.org/S18-1001",
    doi = "10.18653/v1/S18-1001",
    pages = "1--17",
}

@incollection{comrie1986markedness,
author="Comrie, Bernard",
editor="Eckman, Fred R.
and Moravcsik, Edith A.
and Wirth, Jessica R.",
title="Markedness, Grammar, People, and the World",
bookTitle="Markedness",
year="1986",
publisher="Springer US",
address="Boston, MA",
pages="85--106",
isbn="978-1-4757-5718-7",
doi="10.1007/978-1-4757-5718-7_6",
url="https://doi.org/10.1007/978-1-4757-5718-7_6"
}

@article{dressler1985predictiveness,
 ISSN = {00222267, 14697742},
 URL = {http://www.jstor.org/stable/4175791},
 author = {Wolfgang U. Dressler},
 journal = {Journal of Linguistics},
 number = {2},
 pages = {321--337},
 publisher = {Cambridge University Press},
 title = {On the Predictiveness of Natural Morphology},
 urldate = {2024-07-20},
 volume = {21},
 year = {1985}
}

@incollection{levy-etal-2023-comparing,
    title = "Comparing Biases and the Impact of Multilingual Training across Multiple Languages",
    author = "Levy, Sharon  and
      John, Neha  and
      Liu, Ling  and
      Vyas, Yogarshi  and
      Ma, Jie  and
      Fujinuma, Yoshinari  and
      Ballesteros, Miguel  and
      Castelli, Vittorio  and
      Roth, Dan",
    editor = "Bouamor, Houda  and
      Pino, Juan  and
      Bali, Kalika",
    booktitle = "Proceedings of the 2023 Conference on Empirical Methods in Natural Language Processing",
    month = dec,
    year = "2023",
    address = "Singapore",
    publisher = "Association for Computational Linguistics",
    url = "https://aclanthology.org/2023.emnlp-main.634",
    doi = "10.18653/v1/2023.emnlp-main.634",
    pages = "10260--10280",
}

@incollection{ribeiro-etal-2020-beyond,
    title = "Beyond Accuracy: Behavioral Testing of {NLP} Models with {C}heck{L}ist",
    author = "Ribeiro, Marco Tulio  and
      Wu, Tongshuang  and
      Guestrin, Carlos  and
      Singh, Sameer",
    editor = "Jurafsky, Dan  and
      Chai, Joyce  and
      Schluter, Natalie  and
      Tetreault, Joel",
    booktitle = "Proceedings of the 58th Annual Meeting of the Association for Computational Linguistics",
    month = jul,
    year = "2020",
    address = "Online",
    publisher = "Association for Computational Linguistics",
    url = "https://aclanthology.org/2020.acl-main.442",
    doi = "10.18653/v1/2020.acl-main.442",
    pages = "4902--4912",
}

@book{Trubetzkoy1,
    author = {Trubetzkoy, Nikolai Sergeevich},
    title = "Principles of Phonology",
    publisher = "University of California Press",
    year = "1969"
}

@article{Jakobson1,
 ISSN = {00368733, 19467087},
 URL = {http://www.jstor.org/stable/24927429},
 author = {Roman Jakobson},
 journal = {Scientific American},
 number = {3},
 pages = {72--81},
 publisher = {Scientific American, a division of Nature America, Inc.},
 title = {Verbal Communication},
 urldate = {2024-07-18},
 volume = {227},
 year = {1972}
}

@article{Cheryan1,
    author = {Sapna Cheryan and Hazel Rose Markus},
    title = "Masculine defaults: Identifying and mitigating hidden cultural biases.",
    journal = "Psychological Review",
    year = {2020},
    volume = {127},
    number = {6},
    pages = {1022–-1052}
}

@article{Berkel1,
    author = {Laura Van Berkel and Ludwin E. Molina and Sahana Mukherjee},
    title = {Gender asymmetry in the construction of {A}merican national identity},
    journal = "Psychology of Women Quarterly",
    volume = {41},
    pages = {352--367},
    year = {2017}
}

@article{Brekhus1,
    author = "Wayne Brekhus",
    title = "A Sociology of the Unmarked: Redirecting Our Focus",
    journal = "Sociological Theory",
    volume = {16},
    pages = {34--51},
    number = {1},
    year = {2002},
}

@article{Bailey1,
author = {April H. Bailey  and Adina Williams  and Andrei Cimpian },
title = {Based on billions of words on the internet, people=men},
journal = {Science Advances},
volume = {8},
number = {13},
pages = {eabm2463},
year = {2022},
doi = {10.1126/sciadv.abm2463},
URL = {https://www.science.org/doi/abs/10.1126/sciadv.abm2463},
eprint = {https://www.science.org/doi/pdf/10.1126/sciadv.abm2463}}

@incollection{Wolfe1,
author = {Wolfe, Robert and Caliskan, Aylin},
title = {Markedness in Visual Semantic AI},
year = {2022},
isbn = {9781450393522},
publisher = {Association for Computing Machinery},
address = {New York, NY, USA},
url = {https://doi.org/10.1145/3531146.3533183},
doi = {10.1145/3531146.3533183},
booktitle = {Proceedings of the 2022 ACM Conference on Fairness, Accountability, and Transparency},
pages = {1269–1279},
numpages = {11},
location = {Seoul, Republic of Korea},
series = {FAccT '22}
}

@incollection{Wolfe2,
author = {Wolfe, Robert and Caliskan, Aylin},
title = {American == White in Multimodal Language-and-Image AI},
year = {2022},
isbn = {9781450392471},
publisher = {Association for Computing Machinery},
address = {New York, NY, USA},
url = {https://doi.org/10.1145/3514094.3534136},
doi = {10.1145/3514094.3534136},
booktitle = {Proceedings of the 2022 AAAI/ACM Conference on AI, Ethics, and Society},
pages = {800–812},
numpages = {13},
location = {Oxford, United Kingdom},
series = {AIES '22}
}

@inproceedings{zayed-etal-2024-dont,
    title = "Why Don`t Prompt-Based Fairness Metrics Correlate?",
    author = "Zayed, Abdelrahman  and
      Mordido, Goncalo  and
      Baldini, Ioana  and
      Chandar, Sarath",
    editor = "Ku, Lun-Wei  and
      Martins, Andre  and
      Srikumar, Vivek",
    booktitle = "Proceedings of the 62nd Annual Meeting of the Association for Computational Linguistics (Volume 1: Long Papers)",
    month = aug,
    year = "2024",
    address = "Bangkok, Thailand",
    publisher = "Association for Computational Linguistics",
    url = "https://aclanthology.org/2024.acl-long.487/",
    doi = "10.18653/v1/2024.acl-long.487",
    pages = "9002--9019"
}

@incollection{rafailov2023direct,
 author = {Rafailov, Rafael and Sharma, Archit and Mitchell, Eric and Manning, Christopher D and Ermon, Stefano and Finn, Chelsea},
 booktitle = {Advances in Neural Information Processing Systems},
 editor = {A. Oh and T. Naumann and A. Globerson and K. Saenko and M. Hardt and S. Levine},
 pages = {53728--53741},
 publisher = {Curran Associates, Inc.},
 title = {Direct Preference Optimization: Your Language Model is Secretly a Reward Model},
 url = {https://proceedings.neurips.cc/paper_files/paper/2023/file/a85b405ed65c6477a4fe8302b5e06ce7-Paper-Conference.pdf},
 volume = {36},
 year = {2023}
}

@article{PewResearch,
 author  = {Mark Hugo Lopez and Jens Manuel Krogstad and Jeffrey S. Passel},
 date    = {2023-09-05},
 title   = {Who is Hispanic?},
 journal = {Pew Research Center},
 url     = {https://www.pewresearch.org/short-reads/2023/09/05/who-is-hispanic/},
 urldate = {2023-09-05},
 year = 2023
}

@misc{loshchilov2019decoupledweightdecayregularization,
      title={Decoupled Weight Decay Regularization}, 
      author={Ilya Loshchilov and Frank Hutter},
      year={2019},
note = "Preprint at \url{https://arxiv.org/abs/1711.05101}"
}

@inproceedings{cimitan-etal-2024-curation,
    title = "Curation of Benchmark Templates for Measuring Gender Bias in Named Entity Recognition Models",
    author = "Cimitan, Ana  and
      Alves Pinto, Ana  and
      Geierhos, Michaela",
    editor = "Calzolari, Nicoletta  and
      Kan, Min-Yen  and
      Hoste, Veronique  and
      Lenci, Alessandro  and
      Sakti, Sakriani  and
      Xue, Nianwen",
    booktitle = "Proceedings of the 2024 Joint International Conference on Computational Linguistics, Language Resources and Evaluation (LREC-COLING 2024)",
    month = may,
    year = "2024",
    address = "Torino, Italia",
    publisher = "ELRA and ICCL",
    url = "https://aclanthology.org/2024.lrec-main.378/",
    pages = "4238--4246"
}

@inproceedings{Dixon2018,
author = {Dixon, Lucas and Li, John and Sorensen, Jeffrey and Thain, Nithum and Vasserman, Lucy},
title = {Measuring and Mitigating Unintended Bias in Text Classification},
year = {2018},
isbn = {9781450360128},
publisher = {Association for Computing Machinery},
address = {New York, NY, USA},
url = {https://doi.org/10.1145/3278721.3278729},
doi = {10.1145/3278721.3278729},
booktitle = {Proceedings of the 2018 AAAI/ACM Conference on AI, Ethics, and Society},
pages = {67–73},
numpages = {7},
location = {New Orleans, LA, USA},
series = {AIES '18}
}

@inproceedings{huang-etal-2020-reducing,
    title = "Reducing Sentiment Bias in Language Models via Counterfactual Evaluation",
    author = "Huang, Po-Sen  and
      Zhang, Huan  and
      Jiang, Ray  and
      Stanforth, Robert  and
      Welbl, Johannes  and
      Rae, Jack  and
      Maini, Vishal  and
      Yogatama, Dani  and
      Kohli, Pushmeet",
    editor = "Cohn, Trevor  and
      He, Yulan  and
      Liu, Yang",
    booktitle = "Findings of the Association for Computational Linguistics: EMNLP 2020",
    month = nov,
    year = "2020",
    address = "Online",
    publisher = "Association for Computational Linguistics",
    url = "https://aclanthology.org/2020.findings-emnlp.7/",
    doi = "10.18653/v1/2020.findings-emnlp.7",
    pages = "65--83"
}

@inproceedings{martinkova-etal-2023-measuring,
    title = "Measuring Gender Bias in {W}est {S}lavic Language Models",
    author = "Martinkov{\'a}, Sandra  and
      Stanczak, Karolina  and
      Augenstein, Isabelle",
    editor = "Piskorski, Jakub  and
      Marci{\'n}czuk, Micha{\l}  and
      Nakov, Preslav  and
      Ogrodniczuk, Maciej  and
      Pollak, Senja  and
      P{\v{r}}ib{\'a}{\v{n}}, Pavel  and
      Rybak, Piotr  and
      Steinberger, Josef  and
      Yangarber, Roman",
    booktitle = "Proceedings of the 9th Workshop on Slavic Natural Language Processing 2023 (SlavicNLP 2023)",
    month = may,
    year = "2023",
    address = "Dubrovnik, Croatia",
    publisher = "Association for Computational Linguistics",
    url = "https://aclanthology.org/2023.bsnlp-1.17/",
    doi = "10.18653/v1/2023.bsnlp-1.17",
    pages = "146--154"
}

@inproceedings{gupta2024,
    author = {Karan Gupta and Sumegh Roychowdhury and Siva Rajesh Kasa and Santhosh Kasa and Anish Bhanushali and Nikhil Pattisapu and Prasanna Srinivasa Murthy and Alok Chandra},
    title = {How robust are LLMs to in-context majority label bias?},
    booktitle = {AAAI 2024 Workshop on Responsible Language Models},
    year = {2024},
}

@book{Zerubavel1,
	author = {Eviatar Zerubavel},
	publisher = {Princeton University Press},
	title = {Taken for Granted: The Remarkable Power of the Unremarkable},
	year = {2018}}

@misc{gemmateam2024gemmaopenmodelsbased,
      title={Gemma: Open Models Based on Gemini Research and Technology}, 
      author={Gemma and Thomas Mesnard and Cassidy Hardin and Robert Dadashi and Surya Bhupatiraju and Shreya Pathak and Laurent Sifre and Morgane Rivière and Mihir Sanjay Kale and Juliette Love and Pouya Tafti and Léonard Hussenot and Pier Giuseppe Sessa and Aakanksha Chowdhery and Adam Roberts and Aditya Barua and Alex Botev and Alex Castro-Ros and Ambrose Slone and Amélie Héliou and Andrea Tacchetti and Anna Bulanova and Antonia Paterson and Beth Tsai and Bobak Shahriari and Charline Le Lan and Christopher A. Choquette-Choo and Clément Crepy and Daniel Cer and Daphne Ippolito and David Reid and Elena Buchatskaya and Eric Ni and Eric Noland and Geng Yan and George Tucker and George-Christian Muraru and Grigory Rozhdestvenskiy and Henryk Michalewski and Ian Tenney and Ivan Grishchenko and Jacob Austin and James Keeling and Jane Labanowski and Jean-Baptiste Lespiau and Jeff Stanway and Jenny Brennan and Jeremy Chen and Johan Ferret and Justin Chiu and Justin Mao-Jones and Katherine Lee and Kathy Yu and Katie Millican and Lars Lowe Sjoesund and Lisa Lee and Lucas Dixon and Machel Reid and Maciej Mikuła and Mateo Wirth and Michael Sharman and Nikolai Chinaev and Nithum Thain and Olivier Bachem and Oscar Chang and Oscar Wahltinez and Paige Bailey and Paul Michel and Petko Yotov and Rahma Chaabouni and Ramona Comanescu and Reena Jana and Rohan Anil and Ross McIlroy and Ruibo Liu and Ryan Mullins and Samuel L Smith and Sebastian Borgeaud and Sertan Girgin and Sholto Douglas and Shree Pandya and Siamak Shakeri and Soham De and Ted Klimenko and Tom Hennigan and Vlad Feinberg and Wojciech Stokowiec and Yu-hui Chen and Zafarali Ahmed and Zhitao Gong and Tris Warkentin and Ludovic Peran and Minh Giang and Clément Farabet and Oriol Vinyals and Jeff Dean and Koray Kavukcuoglu and Demis Hassabis and Zoubin Ghahramani and Douglas Eck and Joelle Barral and Fernando Pereira and Eli Collins and Armand Joulin and Noah Fiedel and Evan Senter and Alek Andreev and Kathleen Kenealy},
      year={2024},
      eprint={2403.08295},
      archivePrefix={arXiv},
      primaryClass={cs.CL},
      url={https://arxiv.org/abs/2403.08295}, 
}

@misc{qwen2025qwen25technicalreport,
      title={Qwen2.5 Technical Report}, 
      author={Qwen and An Yang and Baosong Yang and Beichen Zhang and Binyuan Hui and Bo Zheng and Bowen Yu and Chengyuan Li and Dayiheng Liu and Fei Huang and Haoran Wei and Huan Lin and Jian Yang and Jianhong Tu and Jianwei Zhang and Jianxin Yang and Jiaxi Yang and Jingren Zhou and Junyang Lin and Kai Dang and Keming Lu and Keqin Bao and Kexin Yang and Le Yu and Mei Li and Mingfeng Xue and Pei Zhang and Qin Zhu and Rui Men and Runji Lin and Tianhao Li and Tianyi Tang and Tingyu Xia and Xingzhang Ren and Xuancheng Ren and Yang Fan and Yang Su and Yichang Zhang and Yu Wan and Yuqiong Liu and Zeyu Cui and Zhenru Zhang and Zihan Qiu},
      year={2025},
      eprint={2412.15115},
      archivePrefix={arXiv},
      primaryClass={cs.CL},
      url={https://arxiv.org/abs/2412.15115}, 
}
\bibliographystyle{tmlr}

\appendix

\section{Fine-Tuning Hyperparameters} \label{app:hyperparameters}

For completeness, we provide the full details of the hyperparameter tuning process used in the fine-tuning experiments. During fine-tuning, early stopping is applied based on validation loss. If no improvement in the loss is observed over a fixed number of steps, then training is stopped. An AdamW optimizer is used with default parameters, except for learning rate (LR) and weight decay \citep{loshchilov2019decoupledweightdecayregularization}. A hyperparameter study was performed to select the best early stopping threshold and LR for all models. For fully fine-tuned models, weight decay was also optimized.

The early stopping threshold was varied between five and seven steps. The learning rate (LR) was chosen from \{1e-3, 3e-4, 1e-4, 3e-5, 1e-5\}, and weight decay, when tuned, was selected from \{1e-3, 1e-4, 1e-5, 1e-6\}. For larger models, LoRA fine-tuning was applied with the rank parameter 8 on every non-embedding layer.

For RoBERTa 125M and 355M and OPT 125M and 350M, 15 training runs were performed, and the five models with the highest accuracy on the SST5 test set were retained. For the larger models, due to resource constraints, five models in total were trained for each model type. Table \ref{param-table} summarizes the optimal hyperparameters selected for each model during fine-tuning.

\begin{table}[ht!]
\caption{Hyperparameters used for model fine-tuning.}
\label{param-table}
\begin{center}
\begin{tabular}{lccc}
\toprule
Model         & Early stop threshold & LR & Weight decay \\ \midrule
RoBERTa-125M  & $7$                      & $1\mathrm{e}{-5}$       & $1\mathrm{e}{-5}$      \\
RoBERTa-355M & $7$                      & $1\mathrm{e}{-5}$       & $1\mathrm{e}{-5}$      \\
OPT-125M      & $7$                      & $1\mathrm{e}{-5}$       & $1\mathrm{e}{-5}$      \\
OPT-350M      & $7$                      & $1\mathrm{e}{-5}$       & $1\mathrm{e}{-3}$        \\
OPT-1.3B      & $5$                      & $1\mathrm{e}{-4}$        & $1\mathrm{e}{-4}$       \\
OPT-6.7B      & $5$                      & $1\mathrm{e}{-4}$        & $1\mathrm{e}{-4}$       \\
OPT-13B       & $5$                      & $1\mathrm{e}{-4}$        & $1\mathrm{e}{-4}$       \\
Llama-2-7B    & $5$                      & $1\mathrm{e}{-4}$         & $1\mathrm{e}{-4}$       \\
Llama-2-13B   & $5$                      & $1\mathrm{e}{-4}$         & $1\mathrm{e}{-4}$      \\
Llama-3-8B    & $5$                      & $1\mathrm{e}{-4}$         & $1\mathrm{e}{-3}$    \\
Mistral-7B    & $5$                      & $3\mathrm{e}{-5}$        & $1\mathrm{e}{-3}$ \\
\bottomrule
\end{tabular}
\end{center}
\end{table}

\section{Prompt Templates and Other Details} \label{app:prompts}

This section includes the templates used in the prompting approach. Each subsection corresponds to a different template. For CoT prompting, inference batches are limited to size  $4$ due to higher computational demands, whereas batch sizes of $16$ are used in other settings.

\subsection{Zero-Shot Prompt Template}

The zero-shot prompt template is displayed below with additional formatting for readability. The component in angled brackets is where each sample to be classified is inserted. The models begin generation at [\emph{LM Generation}].

\noindent \textbf{Text:} \textlangle Text to classify\textrangle \\
\textbf{Question: Is the sentiment of the text negative, neutral, or positive?} \\
\textbf{Answer: The sentiment is} [\emph{LM Generation}]

\subsection{Few-Shot Prompt}

Below is the few-shot template. For the few-shot prompt templates, nine labeled examples are prepended to the prompt, following the template style. The models begin generation at [\emph{LM Generation}].

\noindent\textbf{Text: Example 1 from either SST5 or SemEval} \\
\textbf{Question: What is the sentiment of the text?} \\
\textbf{Answer: Negative.} \vspace{1ex} \\
\textbf{...} \vspace{1ex} \\
\textbf{Text: Example 9 from either SST5 or SemEval} \\
\textbf{Question: What is the sentiment of the text?} \\
\textbf{Answer: Positive.} \vspace{1ex} \\
\textbf{Text:} \textlangle Text to classify\textrangle \\
\textbf{Question: What is the sentiment of the text?} \\
\textbf{Answer:} [\emph{LM Generation}]

\subsection{Zero-Shot CoT Prompt}

Zero-shot CoT uses two prompt templates in sequence. In the first step, the model is provided the text to classify and asked about the corresponding sentiment. The traditional ``trigger'' sentence ``Let’s think step by step'' is used to encourage the model to generate reasoning prior to answering the question. The template appears below.

\noindent\textbf{Text:} \textlangle Text to classify\textrangle \\
\textbf{Question: Is the sentiment of the text negative, neutral, or positive?} \\
\textbf{Reasoning: Let's think step by step.} [\emph{LM Generation}]

In the second step of zero-shot CoT, the reasoning generation is appended to the first prompt along with the answer completion text displayed in the template below. At this stage, the model is expected to generate an answer to be extracted.

\noindent\textbf{Text:} \textlangle Text to classify\textrangle \\
\textbf{Question: Is the sentiment of the text negative, neutral, or positive?} \\
\textbf{Reasoning: Let's think step by step.} \textlangle Generation from previous step\textrangle \\
\textbf{Answer: Therefore, from negative, neutral, or positive, the sentiment is} [\emph{LM Generation}]

\begin{table}[ht!]
\caption{Accuracy statistics on the Amazon dataset for fine-tuning experiments across model types and sizes. Bold numbers indicate the best accuracy achieved within each model family.}
\label{accuracy_table_tune}
\begin{center}
\begin{tabular}{llcc} 
\toprule
Model & Size & Mean Accuracy & Standard Deviation \\
\midrule
\multirow{2}{*}{RoBERTa} & 125M & \textbf{0.635} & 0.036 \\
& 350M & 0.624 & 0.027 \\
\midrule
\multirow{4}{*}{OPT} & 125M & 0.687 & 0.080 \\   
& 350M & 0.692 & 0.039 \\
& 1.3B & \textbf{0.739} & 0.020 \\
& 6.7B & 0.737 & 0.014  \\
\midrule
\multirow{2}{*}{Llama-2} & 7B & 0.513 & 0.089 \\
& 13B & \textbf{0.647} & 0.006 \\
\midrule
\multirow{1}{*}{Llama-3} & 8B & \textbf{0.822} & 0.035 \\
\midrule
\multirow{1}{*}{Mistral} & 7B & \textbf{0.740} & 0.005 \\
\bottomrule
\end{tabular}
\end{center}
\end{table}

\section{Fine-tuning and Prompting Accuracy} \label{appendix_prompt_finetune_tables}

Tables \ref{accuracy_table_tune} and \ref{accuracy_table_prompt} present the average classification accuracy and standard deviation for the fine-tuning and prompting approaches on the Amazon dataset, respectively. Generally, prompt-based classification accuracy is lower than that of fine-tuning. There is also notable variability in model classification accuracy on the template-based probe dataset as a whole, with newer models tending to produce better performance. However, the trends associated with the measured FPR gaps, especially for the White group, are largely consistent, regardless of these variations. Do to this consistency, the differences in model accuracy are unlikely to be a primary contributor to the observed behavior.

\begin{table}[ht!]
\caption{Model accuracy and standard deviation (in parentheses) on the Amazon dataset for prompting experiments across model types. Bold numbers indicate the best accuracy achieved for each model.}
\label{accuracy_table_prompt}
\begin{center}
\begin{tabular}{lcccc}
\toprule
Prompt Type & Zero-shot  & Zero-shot CoT & SemEval 9-shot & SST5 9-shot \\
\midrule
OPT-6.7B & $0.451$ ($0.002$) & -- & $\mathbf{0.482}$ ($0.009$) & 0.433 (0.024) \\
Llama-2-7B & $0.483$ ($0.002$) & $0.492$ ($0.003$) & $\mathbf{0.654}$ ($0.037$) & $0.616$ ($0.028$) \\
Llama-3-8B & $0.600$ ($0.003$) & $0.539$ ($0.001$) & $0.683$ ($0.017$) & $\mathbf{0.716}$ ($0.024$)  \\
Mistral-7B & $0.502$ ($0.003$) & $0.517$ ($0.003$) & $\mathbf{0.700}$ ($0.045$) & $0.682$ ($0.025$)  \\
Gemma-7B & $0.830$ ($0.001$) & $0.777$ ($0.003$) & $\mathbf{0.854}$ ($0.020$) & $0.804$ ($0.031$)  \\
Qwen-2.5-7B & $0.899$ ($0.001$) & $0.823$ ($0.002$) & $\mathbf{0.923}$ ($0.016$) & $0.906$ ($0.001$)  \\
\bottomrule
\end{tabular}
\end{center}
\end{table}

\begin{table}[ht!]
\caption{Models ranked by average gap spans across datasets for Negative- and Positive-Sentiment FPR when fine-tuning. For a given type of FPR, gap spans are computed as the largest difference between any two mean FPR gaps across groups. The larger this span, the greater the difference in Negative- or Positive-Sentiment FPR between groups and the less invariant the model is to group substitution.}
\label{span_table}
\begin{center}
\resizebox{0.99\linewidth}{!}{\begin{tabular}{llC{4.5cm}lC{4.5cm}}
\toprule
Rank & Model & Mean Negative-Sentiment FPR Gap Span & Model & Mean Positive-Sentiment FPR Gap Span \\ 
\midrule
1 & Llama-2-13B & $0.207$ & RoBERTa-355M & $0.154$ \\
2 & RoBERTa-355M & $0.198$ & RoBERTa-125M & $0.152$ \\
3 & Llama-2-7B & $0.144$ & OPT-13B & $0.144$ \\
4 & RoBERTa-125M & $0.126$ & Llama-2-13B & $0.143$ \\
5 & OPT-350M & $0.118$ & Mistral-7B & $0.141$ \\
6 & OPT-1.3B & $0.104$ & OPT-125M & $0.136$ \\
7 & OPT-6.7B & $0.081$ & Llama-2-7B & $0.133$ \\
8 & OPT-13B & $0.056$ & OPT-350M & $0.132$ \\
9 & OPT-125M & $0.039$ & OPT-1.3B & $0.128$ \\
10 & Mistral-7B & $0.032$ & OPT-6.7B & $\mathbf{0.089}$ \\
11 & Llama-3-8B & $\mathbf{0.020}$ & Llama-3-8B & $\mathbf{0.089}$ \\
\bottomrule
\end{tabular}}
\end{center}
\end{table}

\section{Gap Differences Across Models} \label{appendix_span_table}

For each of the fine-tuned models, across the different datasets, an FPR-gap span is calculated. For a given type of FPR, Negative- or Positive-Sentiment gap spans are computed as the largest difference between any two mean FPR gaps for the groups. This quantifies how large the particular FPR disparities for a given model and dataset are between groups. The larger this span, the greater the difference in Negative- or Positive-Sentiment FPR between groups and the less invariant the model is to overall group substitution. Table \ref{span_table} displays the FPR gap spans for each model, averaged over the three datasets. In computing the spans, the gap for the White group is part of the span extremes 58\% of the time for Negative-Sentiment FPR and 100\% of the time for Positive-Sentiment FPR. That is, the gap computed for the White group often constitutes one of the largest gap magnitudes.

From the table, it is clear that the RoBERTa and Llama-2 models have consistently large spans for both types of FPR gap. On the other hand, Llama-3-8B, the most recent model studied with fine-tuning, has the smallest average gap spans in both categories. Another recent model, Mistral-7B, demonstrates a small average Negative-Sentiment FPR gap span, suggesting that more recent LLMs may be slightly less affected by issues with the template-based probes. It is interesting to note that the distribution of spans for Positive-Sentiment FPR gaps are more uniformly distributed between models than the Negative-Sentiment counterpart.

\begin{table}[ht!]
    \caption{Statistics for the three template-based probing datasets, Amazon, Regard, and NS-Prompts. The label counts and distributions are reported for each racial group across datasets. The label distributions are constant between groups, and thus reported below the label totals. Note that all examples from the NS-Prompts dataset have a neutral label by construction.}
    \label{probing_dataset_stats}
    \begin{center}
    \begin{tabular}{llccccccc}
    \toprule
    & & \multicolumn{3}{c}{Label Counts} & & \\
    \cmidrule(lr){3-5}
    Dataset & Race & Negative & Neutral & Positive & Total & Fraction \\
    \midrule
    \multirow{8}{*}{Amazon} & African American & $200$ & $200$ & $200$ & $600$ & $0.182$ \\
    & American Indian & $300$ & $300$ & $300$ & $900$ & $0.273$ \\
    & Asian & $100$ & $100$ & $100$ & $300$ & $0.091$ \\
    & Hispanic & $200$ & $200$ & $200$ & $600$ & $0.182$ \\
    & Pacific Islander & $200$ & $200$ & $200$ & $600$ & $0.182$ \\
    & White & $100$ & $100$ & $100$ & $300$ & $0.091$ \\
    \cmidrule(lr){2-7}
    & Total & $1100$ & $1100$ & $1100$ & $3300$ & \\
    & Distribution & $0.333$ & $0.333$ & $0.333$ & & \\
    \midrule
    \multirow{8}{*}{Regard} & African American & $440$ & $220$ & $350$ & $1010$ & $0.182$  \\
    & American Indian & $660$ & $330$ & $525$ & $1515$ & $0.273$ \\
    & Asian & $220$ & $110$ & $175$ & $505$ & $0.091$ \\
    & Hispanic & $440$ & $220$ & $350$ & $1010$ & $0.182$ \\
    & Pacific Islander & $440$ & $220$ & $440$ & $1010$ & $0.182$ \\
    & White & $220$ & $110$ & $175$ & $505$ & $0.091$ \\
    \cmidrule(lr){2-7}
    & Total & $2420$ & $1210$ & $1925$ & $5555$ & \\
    & Distribution & $0.437$ & $0.218$ & $0.347$ & & \\
    \midrule
    \multirow{8}{*}{NS-Prompts} & African American & $0$ & $8880$ & $0$ & $8880$ & $0.182$ \\
    & American Indian & $0$ & $13320$ & $0$ & $13320$ & $0.273$ \\
    & Asian & $0$ & $4440$ & $0$ & $4440$ & $0.091$ \\
    & Hispanic & $0$ & $8880$ & $0$ & $8880$ & $0.182$ \\
    & Pacific Islander & $0$ & $8880$ & $0$ & $8880$ & $0.182$ \\
    & White & $0$ & $4440$ & $0$ & $4440$ & $0.091$ \\
    \cmidrule(lr){2-7}
    & Total & $0$ & $48840$ & $0$ & $48840$ & \\
    & Distribution & $0.0$ & $1.0$ & $0.0$ & & \\
    \bottomrule
    \end{tabular}
    \end{center}
\end{table}

\section{Template-Based Dataset Statistics} \label{template_dataset_stats}

In this appendix, some statistics and distribution information about the template-based probing datasets described in Section \ref{template_datasets} are presented. Table \ref{probing_dataset_stats} summarizes the label counts across datasets and broken down by racial groups. Note that the label distributions are constant between groups. For example, the label distribution of $(0.333, 0.333, 0.333)$ holds for each protected group as well as the aggregated label distribution. Further, it is important to recall that the models are not trained on these datasets and only perform classification inference to produce the results of Section \ref{results}. For the Amazon dataset, the labels are balanced over the entire dataset and within each individual group. The distribution for the Regard dataset is also fairly balanced with somewhat fewer neutral examples. Finally, as discussed in Section \ref{ns_prompts_dataset}, all examples for the NS-Prompts dataset have a neutral label.

\section{Supporting Results for Unmarked Groups in Other Sensitive Attributes} \label{sexuality_and_gender_results}

This work primarily focuses on unexpected measurements with respect to the sensitive attribute of race. However, the conjecture that markedness causes, or at least contributes to, the observed irregularities when using template-based probes would, in theory, generalize beyond racial demographics. For example, as discussed in Section \ref{markedness_conjecture}, several previous studies have noted that gender-based markedness is also prevalent in web data and influences model behaviors. In such settings, male gender represents the unmarked default. Furthermore, linguistic markedness extends to sexuality, with heterosexuality comprising the predominant unmarked demographic \citep{Zerubavel1}. 

With this in mind, two additional experiments are conducted using the prompting approaches described in Section \ref{prompting_setup} and the Amazon dataset \citep{czarnowska2021quantifying}. The Amazon dataset includes templates for sensitive attributes beyond race, including sexuality and gender, with the same structure described in Section \ref{amazon_dataset}. Prompt-based classification is applied to these templates, and the same FPR gaps are measured across various protected groups within the sensitive attributes. 

\begin{figure}[ht!]
\centering
\includegraphics[width=0.99\textwidth]{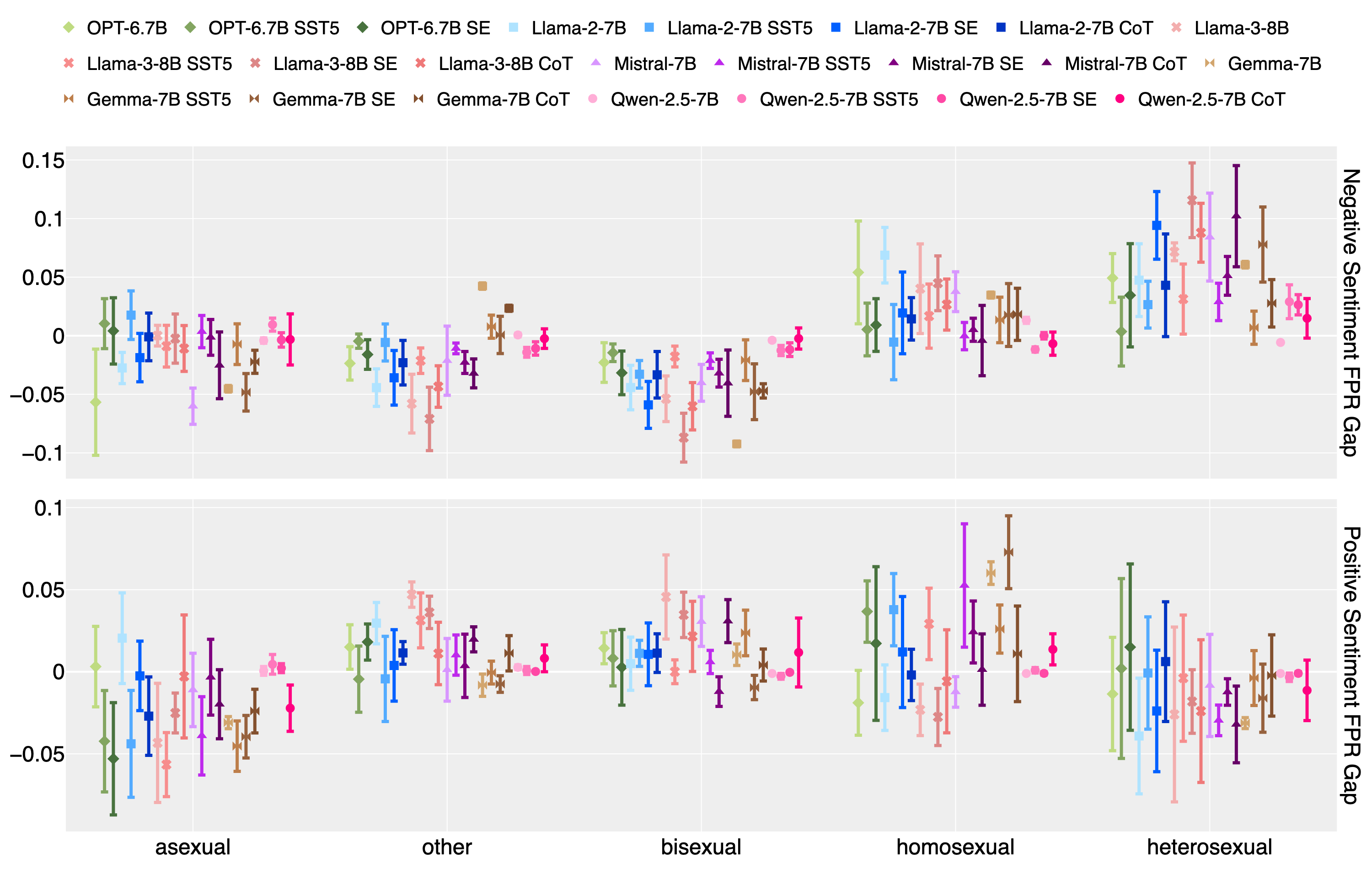}
\caption{Negative- and Positive-Sentiment FPR gaps for protected group variations within the sensitive attribute of Sexuality as measured by the Amazon dataset. In the legend, model names without a suffix indicate zero-shot prompting. SST5 and SE indicate 9-shot prompts with examples drawn from the SST5 and SemEval datasets, respectively.}
\label{amazon_sexuality_gaps_fig} 
\end{figure}

\begin{figure}[ht!]
\centering
\includegraphics[width=0.99\textwidth]{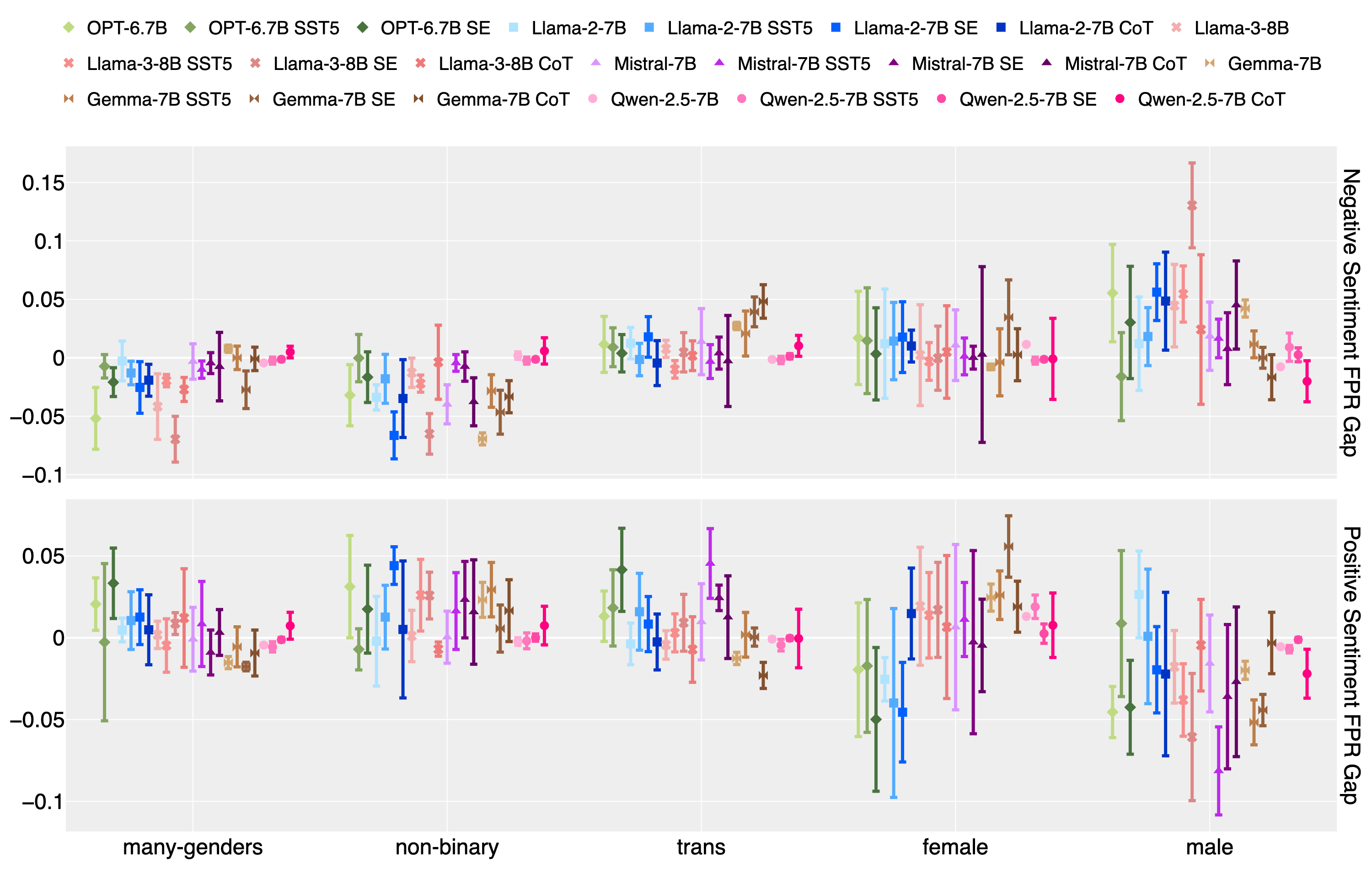}
\caption{Negative- and Positive-Sentiment FPR gaps for protected group variations within the sensitive attribute of Gender as measured by the Amazon dataset. In the legend, model names without a suffix indicate zero-shot prompting. SST5 and SE indicate 9-shot prompts with examples drawn from the SST5 and SemEval datasets, respectively.}
\label{amazin_gender_gaps_fig} 
\end{figure}

The results are shown in Figures \ref{amazon_sexuality_gaps_fig} and \ref{amazin_gender_gaps_fig}. For sexuality, the traditionally unmarked group, heterosexual, reflects a similar FPR-gap pattern to the White group in the main results. That is, positive, and often statistically significant, Negative-Sentiment FPR gaps and negative Positive-Sentiment FPR gaps are found. Moreover, these patterns correlate to traditionally underprivileged groups in homosexual and asexual orientation for the respective FPR-gap types. When considering gender, the templates associated with male gender also produce positive and negative gaps for Negative- and Positive-Sentiment FPR, respectively, though the frequency of statistical significance is slightly reduced. 

As noted in Section \ref{markedness_conjecture}, additional experimentation is required to confirm that markedness is a driving factor for the results presented in this paper. Nonetheless, the empirical results in this section reinforce the observation that template-based probes produce imperfect measures of bias in LLMs and that these imperfections appear to affect measurements associated with traditionally unmarked groups, even beyond of the sensitive attribute of race. 

\end{document}